\documentclass[10pt,twocolumn,letterpaper]{article}

\usepackage{cvpr}              
\usepackage[T1]{fontenc}      
\usepackage{url}            
\usepackage{booktabs}       
\usepackage{amsfonts}       
\usepackage{nicefrac}       
\usepackage{microtype}      
\usepackage{multirow}
\usepackage{amsmath}
\usepackage{graphicx}
\usepackage{svg}
\usepackage{scalefnt}
\usepackage{chngcntr}
\usepackage{amssymb}
\usepackage{caption}
\usepackage{tcolorbox}
\tcbuselibrary{breakable} 
\usepackage{lipsum}         
\usepackage[table,xcdraw]{xcolor}
\usepackage{siunitx}   
\usepackage{colortbl}
\usepackage{adjustbox}
\usepackage{array}     
\usepackage{longtable} 
\usepackage{calc}      
\usepackage{titletoc} 
\usepackage{tabularx}
\usepackage{ragged2e} 
\usepackage{tabularray}
\usepackage{marvosym}  

\definecolor{lightgray}{HTML}{EFEFEF}
\definecolor{RowColorA}{rgb}{0.95, 0.95, 0.98} 
\definecolor{RowColorB}{rgb}{0.92, 0.95, 0.98} 
\definecolor{RowColorC}{rgb}{0.90, 0.93, 0.96} 

\tcbuselibrary{skins} 

\definecolor{lightviolet}{rgb}{0.984, 0.937, 0.976}

\definecolor{cvprblue}{rgb}{0.21,0.49,0.74}
\usepackage[pagebackref,breaklinks,colorlinks,allcolors=cvprblue]{hyperref}

\colorlet{Mycolor1}{green!100}

\def\paperID{} 
\def\confName{CVPR}
\def\confYear{2026}

\title{AutoGUI-v2: A Comprehensive Multi-Modal GUI Functionality Understanding Benchmark}

\author{
  Hongxin Li$^{*1,2,3}$\quad 
  Xiping Wang$^{*1,2,3}$ \quad
  Jingran Su$^{5}$ \quad 
  Zheng Ju$^{1,2,3}$ \quad
  Yuntao Chen$^{4}$ \quad 
  Qing Li$^{5}$\\
  Zhaoxiang Zhang$^{1,2,3,6}$\textsuperscript{\Letter} \quad
  \\ 
$^1$University of Chinese Academy of Sciences (UCAS) \\
$^2$New Laboratory of Pattern Recognition (NLPR), CASIA \\
$^3$State Key Laboratory of Multimodal Artificial Intelligence Systems (MAIS), CASIA \\
$^4$Hong Kong Institute of Science \& Innovation, CASIA \\
$^5$PolyU \quad
$^6$Shanghai Artificial Intelligence Laboratory \quad
 \\
  \small{Code: \url{https://github.com/ZJULiHongxin/AutoGUI-v2}}
}

\begin{document}

\twocolumn[{%
\vspace{-2em}
\maketitle%

{
    \centering
    \includegraphics[width=1\textwidth]{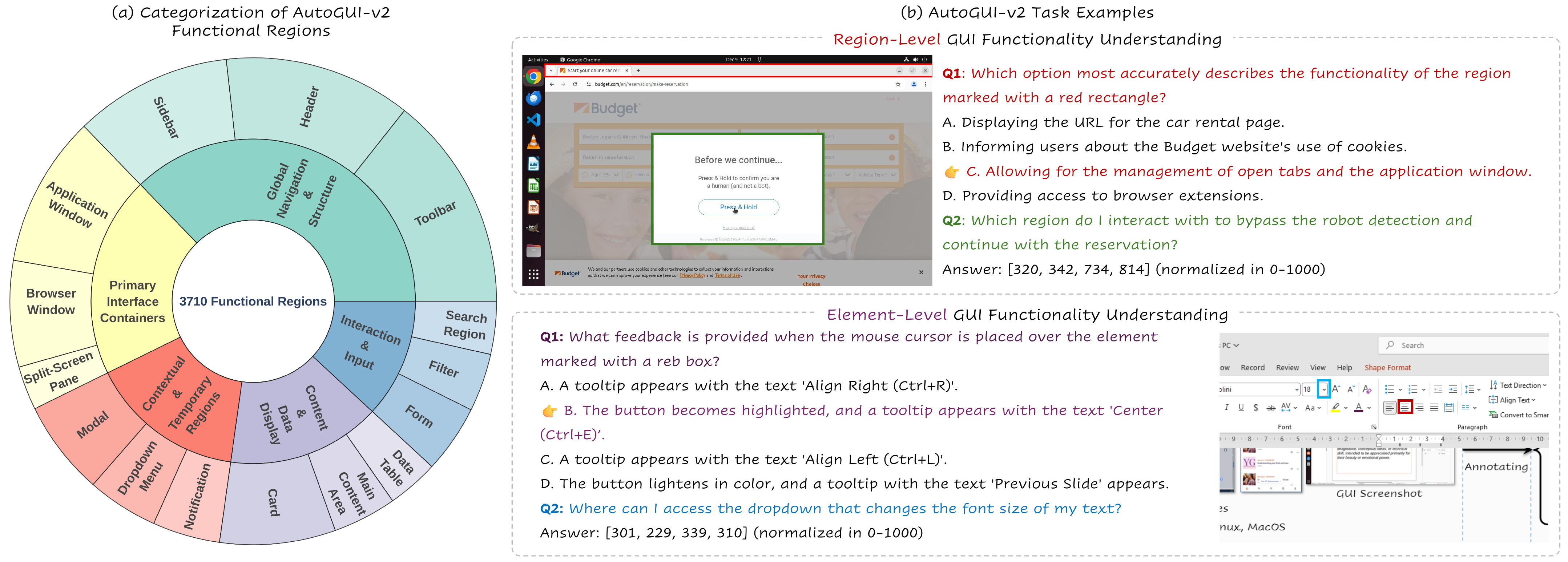}
    \captionof{figure}{\textbf{AutoGUI-v2 benchmark overview.} (a) Region types covered. (b) Representative functionality understanding tasks.}
    \label{fig:teaser}
    \vspace{1.5em}
}%
}]

\newcommand{\myfootnote}[1]{%
  \begingroup
  \renewcommand\thefootnote{}\footnotetext{#1}%
  \endgroup
}

\myfootnote{\textsuperscript{*} Co-authors. \textsuperscript{\Letter} Corresponding author.}

\begin{abstract}
Autonomous agents capable of navigating Graphical User Interfaces (GUIs) hold the potential to revolutionize digital productivity. However, achieving true digital autonomy extends beyond reactive element matching; it necessitates a predictive mental model of interface dynamics and the ability to foresee the ``digital world state'' resulting from interactions. Despite the perceptual capabilities of modern Vision-Language Models (VLMs), existing benchmarks remain bifurcated—focusing either on black-box task completion or static, shallow grounding—thereby failing to assess whether agents truly comprehend the implicit functionality and transition logic of GUIs. To bridge this gap, we introduce \textbf{AutoGUI-v2}, a comprehensive benchmark designed to evaluate deep GUI functionality understanding and interaction outcome prediction. We construct the benchmark using a novel VLM-human collaborative pipeline that recursively parses multi-platform screenshots into hierarchical functional regions to generate diverse evaluation tasks. Providing 2,753 tasks across six operating systems, AutoGUI-v2 rigorously tests agents on region and element-level semantics, grounding, and dynamic state prediction. Our evaluation reveals a striking dichotomy in VLMs: while open-source models fine-tuned on agent data (e.g., Qwen3-VL) excel at functional grounding, commercial models (e.g., Gemini-2.5-Pro-Thinking) dominate in functionality captioning. Crucially, all models struggle with complex interaction logic of uncommon actions, highlighting that deep functional understanding remains a significant hurdle. By systematically measuring these foundational capabilities, AutoGUI-v2 offers a new lens for advancing the next generation of GUI agents.
\end{abstract}    
\section{Introduction}
\label{sec:intro}

The development of autonomous agents capable of navigating digital interfaces represents a transformative frontier in artificial intelligence, promising to redefine human-computer interaction~\cite{sheetcopilot,cheng2024seeclick, UIPro, mpgui, osatlas, uitars, aguvis, hong2023cogagent, uground, aria-ui,showui,SpiritSight}.
While recent Vision-Language Models (VLMs)~\cite{liu2023llava,llamav,qwen2vl,qwen3vl,chen2023internvl} have empowered agents to interact with GUIs across diverse platforms, effective digital autonomy requires more than reactive element grounding~\cite{li2025autoguiscalingguigrounding,shi2025guiknowledgebenchrevealing,zhang2025tongui}.
It demands a profound understanding of interface dynamics, the implicit functionality of regions, and the ability to predict the ``digital world state'' resulting from an action~\cite{li2025autoguiscalingguigrounding}.
Such predictive reasoning is a hallmark of human competence and is essential for agents to generalize across diverse scenarios~\cite{drivewm,richens2025general,NWM}.

\begin{figure}[tp]
    \centering
    \includegraphics[width=0.9\linewidth]{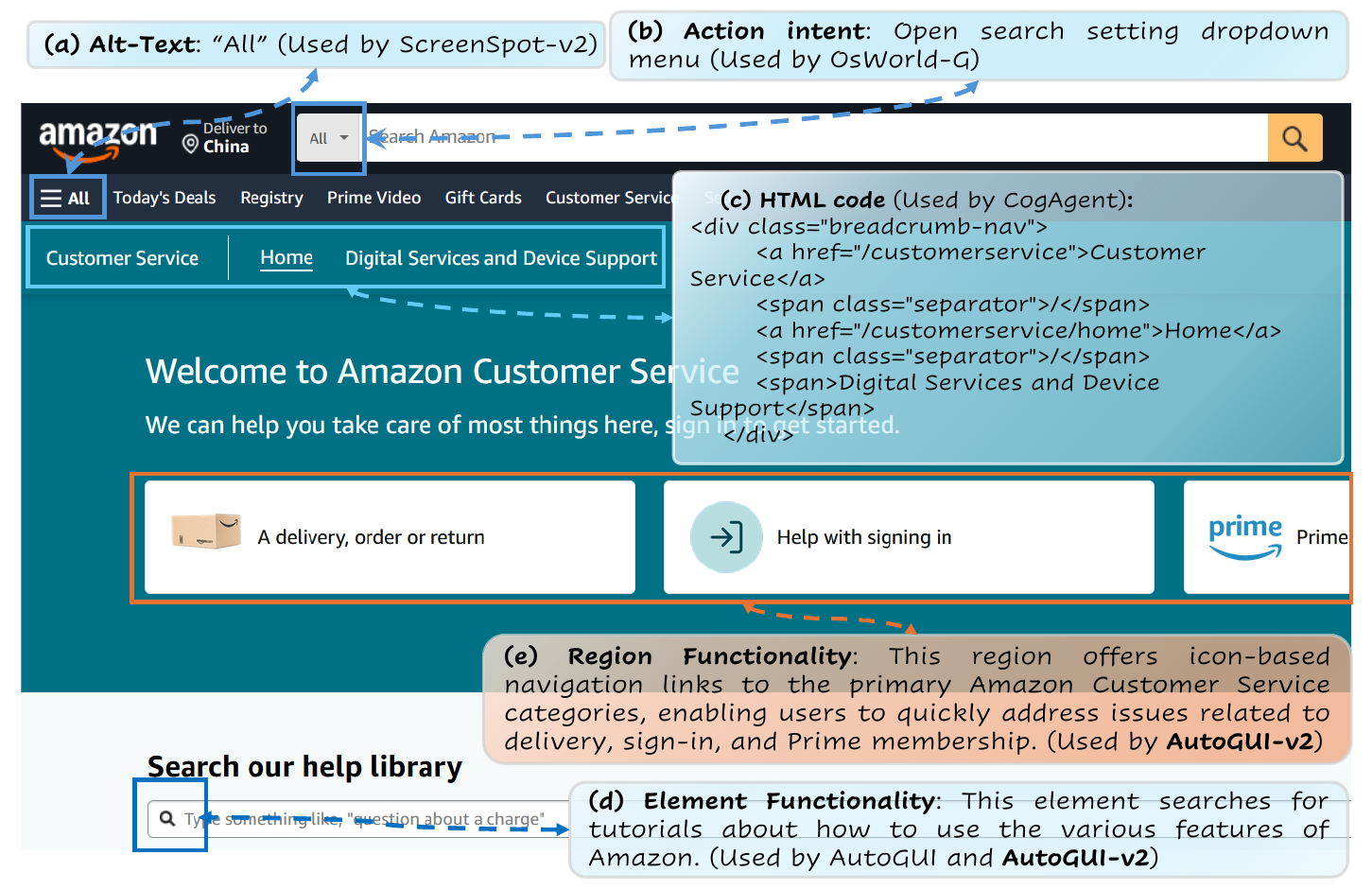}
    \caption{AutoGUI-v2 provides rich functional semantics for both GUI elements and regions compared with existing benchmarks.}
    \label{fig: retype comp}
\end{figure}

Despite this requirement, existing benchmarks largely overlook this deeper functional understanding, falling into two primary categories.
The first, \emph{task-completion benchmarks}~\cite{AndroidWorld, rawles2023android, AITZ, OSWorld, deng2024mind2web, GUIOdyssey}, treat the agent as a black box, assessing success rates without diagnosing how the agent interprets GUI functionality. The second, \emph{grounding benchmarks}~\cite{cheng2024seeclick, screenspotpro, li2025autoguiscalingguigrounding, liu2024visualwebbench, osworldg,uivision}, offer fine-grained evaluation but are limited to simplistic element localization.
These benchmarks typically rely on brief appearance descriptions~\cite{Li2020WidgetCG,cheng2024seeclick}, alt-texts~\cite{liu2024visualwebbench}, or action intents~\cite{osworldg} and fail to test the understanding of transition logic or GUI context, as illustrated in Fig.~\ref{fig: retype comp}.
While the recent AutoGUI~\cite{li2025autoguiscalingguigrounding} introduced functionality understanding evaluation, it remains constrained by platform diversity and lacks tasks probing transition dynamics.
Consequently, a critical gap remains: the field lacks a benchmark that measures an agent's foundational model of GUI functionality and interaction outcomes.

To bridge this gap, we introduce \textbf{AutoGUI-v2}, a comprehensive benchmark designed to evaluate a VLM-based GUI agent's deep understanding of both element-level and region-level functionality.
AutoGUI-v2 is constructed using a novel VLM-human collaborative pipeline. We utilize \textbf{Gemini-2.5-Pro-Thinking}~\cite{gemini} to recursively divide multi-platform screenshots into hierarchical functional regions, which are subsequently verified by a VLM-based scorer and refined via manual annotation to ensure high precision.
Leveraging these hierarchical divisions, we generate challenging tasks that require agents to localize targets based on functional descriptions and predict interaction outcomes.
Complementing this, we employ \textbf{OmniParser-v2}~\cite{OmniParser} to assist in element-level grounding and captioning task generation.
In total, AutoGUI-v2 comprises \textbf{2,753} evaluation tasks across six operating systems, serving as a rigorous testbed for GUI functionality comprehension.

Evaluation on AutoGUI-v2 reveals a striking divergence in capabilities: Open-source models (e.g., Qwen3-VL) unexpectedly outperform leading commercial models (e.g., Gemini-2.5-Pro-Thinking) at functionality-oriented grounding—the task of localizing ``where'' a function is. Conversely, these same commercial models dominate the functionality captioning task, or reasoning ``what'' a region/element does.
Furthermore, our analysis pinpoints why models fail. Performance plummets for irregular region types and  complex interactions, indicating the models rely on overt cues and fail to grasp implicit functionality. Finally, these models are consistently tricked by ``hard'' plausible distractors (i.e., visually similar but functionally distinct candidates) in our functionality captioning tasks, proving their failures stem from a lack of context-aware functionality understanding, not random error.

Our contributions are summarized as follows:
\begin{itemize}
\item We introduce a scalable pipeline for detecting and annotating hierarchical functional regions across multi-platform GUIs, providing a valuable resource of region-level annotations to the community.
\item We propose \textbf{AutoGUI-v2}, a large-scale benchmark evaluating deep GUI functionality understanding at both region and element levels, moving beyond simple grounding to assessing the comprehension of GUI dynamics.
\item We provide a comprehensive analysis of leading VLMs, offering unprecedented insights into their limitations in understanding GUI functionality and state transitions.

\end{itemize}

\section{Related Works}
\label{sec:related works}

\begin{table}[]
\caption{Comparing our AutoGUI-v2 dataset with existing popular GUI understanding benchmarks.}
\label{tab:bmk comparison}
\resizebox{\columnwidth}{!}{%
\begin{tabular}{@{}ccccccc@{}}
\toprule
Benchmark           & \begin{tabular}[c]{@{}c@{}}\#Operating\\ Systems\end{tabular}& \begin{tabular}[c]{@{}c@{}}Functionality\\ Annotation\\ Type\end{tabular} & Max Res. & Target Type & \#Tasks \\ \midrule
ScreenSpot~\cite{cheng2024seeclick}          & 5        & N/A                                                                       &   $1280 \times 720$    &    Element    & 1272    \\
ScreenSpot-Pro~\cite{screenspotpro}      & 3        & N/A                                                                       &   $3840 \times 2160$    &    Element     & 1581    \\
VisualWebBench~\cite{liu2024visualwebbench}      & 1        & N/A                                                                       &   $3840 \times 2160$    &    Element     & 1581    \\

OSWorld-G~\cite{xie2025scalingcomputerusegroundinguser}           &  1       & N/A                                                                       &    $1920 \times 1080$    &   Element     & 510     \\

UI-Vision~\cite{uivision}           &  1       & N/A                                                                       &    $1920 \times 1080$    &   Element \& Region    & 8227     \\
MMBench-GUI~\cite{mmbenchgui}         & 6      & N/A                                                                       &   $3840 \times 2160$      &    Element    & 8000+   \\
GUI-Knowledge-Bench~\cite{shi2025guiknowledgebenchrevealing} & 6       & Brief                                                                     &   $3840 \times 2160$     &   Element    & 3483    \\
AutoGUI~\cite{li2025autoguiscalingguigrounding}             & 2       & Contextual                                                                &  $1280 \times 720$    &   Element      & 2000    \\
AutoGUI-v2 (ours)          & 6      & Contextual                                                                &    $3840 \times 2160$     & Element \& Region  &   2753  \\ \bottomrule
\end{tabular}%
}
\end{table}
\begin{figure*}[th]
  \centering
   \includegraphics[width=1.0\linewidth]{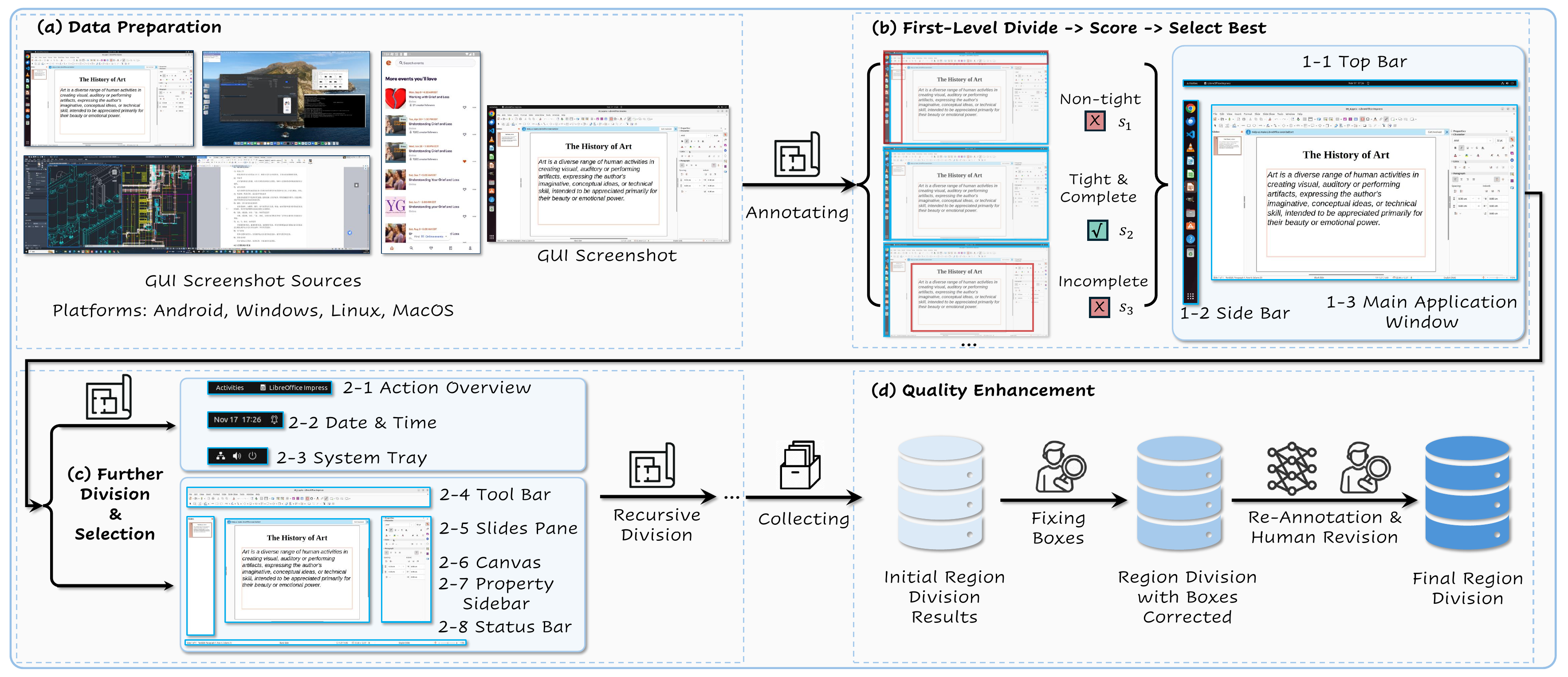}

   \caption{Overview the AutoGUI-v2 Annotation Pipeline. (a) The process begins by sourcing diverse, multi-platform GUI screenshots. (b) A VLM (Gemini-2.5-Pro-Thinking) proposes an initial, first-level decomposition into functional regions, which are automatically scored for quality. (c) This ``divide-and-verify'' process is applied recursively, progressively breaking down complex regions into their granular, non-dividable components. (d) In the final stage, human experts provide pixel-perfect refinement of all bounding boxes, enabling a final VLM pass to generate the high-fidelity functionality descriptions.}
   \label{fig:autoguiv2 anno pipeline}
\end{figure*}
\subsection{GUI Benchmarks}

The evolution of GUI benchmarking has progressed through three distinct paradigms. The foundation lies in \emph{grounding capability evaluation}, which tests an agent's ability to localize UI elements from natural language instructions; this was pioneered by ScreenSpot~\cite{cheng2024seeclick} and expanded by high-resolution and multi-application benchmarks such as ScreenSpot Pro~\cite{screenspotpro}, UI-Vision~\cite{uivision}, and OSWorld-G~\cite{osworldg}. Subsequently, \emph{offline agent evaluation} emerged to assess agents on static environment snapshots, exemplified by Mind2Web~\cite{deng2024mind2web} for web, GUI Odyssey~\cite{GUIOdyssey} for mobile, and the cross-platform GUICourse~\cite{GUICourse}. Most recently, the field has advanced to \emph{online agent evaluation} in dynamic environments via tools like OSWorld~\cite{OSWorld}, AndroidWorld~\cite{AndroidWorld}, AndroidLab~\cite{AndroidLab}, MacOSWorld~\cite{macOSWorld}, and WAA~\cite{windowsarena}.
However, these benchmarks typically overlook the understanding of interaction-induced state changes, failing to measure if agents grasp essential GUI dynamics. AutoGUI-v2 fills this gap by benchmarking comprehensive GUI functionality understanding. Unlike the concurrent GUI Knowledge Bench~\cite{shi2025guiknowledgebenchrevealing}, which addresses interaction outcomes but lacks region-level granularity and sufficient context for element localization, our approach ensures a robust evaluation of underlying GUI functionality.

\subsection{Multi-Modal GUI Agents}

The development of multimodal GUI agents has shifted from API-dependent systems like AppAgent~\cite{zhang2023appagent}, SheetCopilot~\cite{sheetcopilot}, and OmniParser~\cite{OmniParser} to scalable, end-to-end approaches—such as CogAgent~\cite{hong2023cogagent}, ShowUI~\cite{showui}, UGround~\cite{uground}, SpiritSight~\cite{SpiritSight}, Aguvis~\cite{aguvis}, Jedi~\cite{xie2025scalingcomputerusegroundinguser}, OS-Atlas~\cite{osatlas}, GUI-Xplore~\cite{guiexplore}, and UI-Tars~\cite{uitars}. Recently, these visual methods have been augmented by reinforcement learning paradigms~\cite{lu2025ui,liu2025infigui,tang2025guig2gaussianrewardmodeling}, which enhance generalization and reduce data reliance in complex environments.
However, while agent architectures have advanced, evaluation methodologies remain fixated on task execution rates, neglecting the foundational understanding of GUI dynamics. Our AutoGUI-v2 addresses this gap by probing the underlying ``digital state prediction'' useful for agent ability~\cite{richens2025general}.

\section{AutoGUI-v2 Construction}
\label{sec:method}

\begin{figure*}[t]
    \centering
    \includegraphics[width=1.0\linewidth]{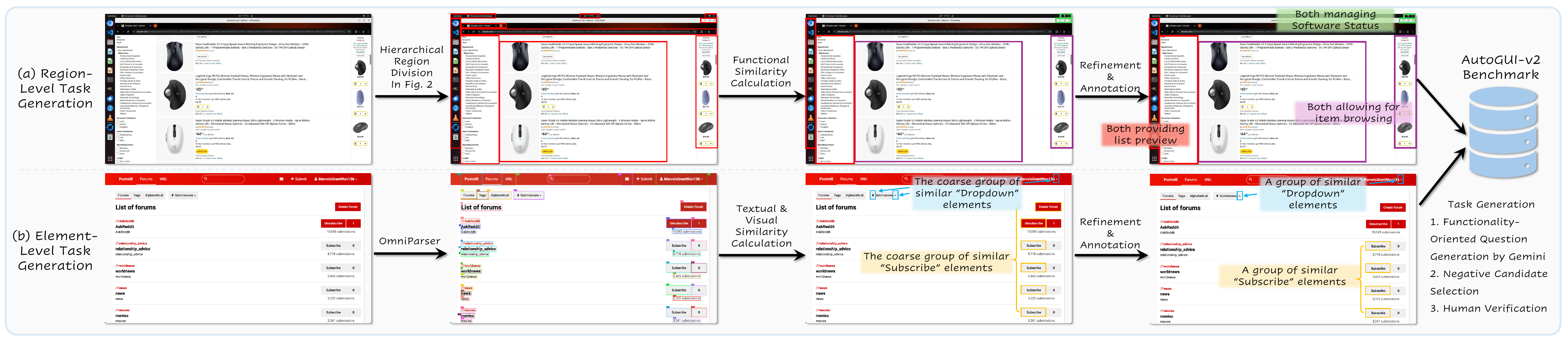}
    \caption{The pipeline of generating AutoGUI-v2 benchmark tasks introduced in Sec.~\ref{sec:region task gen} and Sec.~\ref{sec:element task gen}.}
    \label{fig: task gen}
\end{figure*}

This section details the AutoGUI-v2 annotation pipeline (Fig.~\ref{fig:autoguiv2 anno pipeline}) and task generation methodology (Fig.~\ref{fig: task gen}).

\subsection{Definitions}

A \textbf{functional region} is defined as a high-level grouping of individual UI elements (such as buttons, input fields, links, navigation elements, or icons) that collectively serve a unified purpose. These regions are organized hierarchically, where parent regions are recursively decomposed into granular child units (Fig.~\ref{fig:autoguiv2 anno pipeline}). The region types are exemplified in Fig.~\ref{fig:teaser} and detailed in Sec.~\ref{sec:supp:region type cls} of Appendix.

At the element level, we focus on \textbf{visually similar elements with distinct functionality}. These visually analogous yet functionally distinct components force models to rely on context rather than appearance alone.

\subsection{Data Sources}

We aggregate GUI screenshots from diverse sources, including ScreenSpot-Pro~\cite{screenspotpro}, OSWorld-G~\cite{osworldg}, MMBench-GUI~\cite{mmbenchgui}, AgentNet~\cite{wang2026opencua}, and AMEX~\cite{amex}. This multi-source strategy ensures robust coverage across desktop, web, and mobile domains (Details in Appendix).

\subsection{Region-Level Functionality Understanding}

To construct high-quality, hierarchical annotations, we employ a three-stage human-in-the-loop pipeline.

\subsubsection{Hierarchical Region Annotation Pipeline}

\noindent \textbf{Stage 1: Automated Proposal \& Verification.} Drawing inspiration from the AutoGUI pipeline~\cite{li2025autoguiscalingguigrounding}, we leverage \textit{Gemini-2.5-Pro-Thinking}~\cite{gemini} to automatically annotate functional regions through a multi-step process. Given a region image $\mathbf{I}_{region} \in \mathbb{R}^{H \times W \times 3}$, the VLM identifies $K$ first-level sub-regions, predicting each region's functionality $\{F_i\}$ with instance-specific contextual details, layout description $\{D_i\}$, normalized bounding boxes $\{B_i\}$, and a divisibility flag $\{d_i\}$. This process is formulated as:
\begin{equation}
    \{F_i, D_i, B_i, d_i\}_{i=1}^K = \text{VLM}(\mathbf{I}_{region}, p_{anno})
\end{equation}
where $p_{anno}$ is the annotation prompt shown in Appendix.
Afterwards, this pipeline recursively divides the regions with $d_i=\text{True}$ until all regions are no longer dividable.

To mitigate bounding box errors (examples in Appendix), we incorporate a \textbf{refinement mechanism}. A separate VLM instance evaluates the cropped sub-region $\mathbf{I}_i$ against the root screenshot $\mathbf{I}_{root}$ on two metrics: (1) \textbf{Completeness} $s^{comp}_i \in \{0, 1\}$, checking for full visibility, and \textbf{Boundedness} $s^{bound}_i \in \{0, 1\}$, ensuring tight framing.
\begin{equation}
    s^{comp}_i, s^{bound}_i = \text{VLM}(\mathbf{I}_{root}, \mathbf{I}_{i}, p_{scoring})
\end{equation}
where $p_{scoring}$ is the scoring prompt shown in Appendix.

For an empirically better tradeoff of cost and quality, if a proposal obtains the average scores $\Sigma_{i=1}^k s^{comp}_i / K >= 0.8 $ and $\Sigma_{i=1}^k s^{bound}_i / K >= 0.8$, the proposal is accepted; otherwise, the model regenerates up to $N=3$ iterations. Note that this mechanism is not intended for perfect division, but to provide accurate initial divisions for human correction in the next stage (More  details in the appendix).

\noindent \textbf{Stage 2: Human-in-the-Loop Bounding Correction.} To achieve benchmark-grade precision, the authors refine the automated proposals using a custom web interface (details in Sec.~\ref{sec:supp:reanno details}). This tool enables annotators to traverse region trees and manually adjust bounding boxes $B_i$ to tightly encompass the regions, correcting strictly for spatial accuracy.

\noindent \textbf{Stage 3: Functionality Re-annotation.} Since manual spatial corrections alter the visual content of the crops, we perform a final automated pass to align the semantic annotations with the refined regions. The corrected crop $\mathbf{I}^{fixed}_i$ is fed back into the VLM to generate updated functionality $F_i^{fixed}$ and descriptions $D_i^{fixed}$:
\begin{equation}
    F_i^{fixed}, D_i^{fixed} = \text{VLM}(\mathbf{I}_{root}, \mathbf{I}_{i}^{fixed}, p_{reanno})
\end{equation}
Finally, human annotators validate these re-annotations against our quality criteria (i.e., Contextuality, High-Level Purpose, and Non-Hallucination) explained in Sec.~\ref{sec:supp:reanno details}.

\subsubsection{Task Generation}
We generate evaluation tasks by identifying groups of visually similar regions that serve distinct functions.
\label{sec:region task gen}

\noindent \textbf{Stage 1: Semantic Clustering.} We initialize candidate groups by computing a cosine similarity matrix on the embeddings of region descriptions ($D_i$), generated via Qwen3-Embedding~\cite{qwen3embedding}. Regions with similarity scores $s_{sem} > th_{sem}$ are clustered, effectively grouping regions that share visual attributes (e.g., layout) as described in text space.

\noindent \textbf{Stage 2: Group Verification and Refinement.} A VLM (i.e. Gemini-2.5-Pro-Thinking) verifies these initial clusters within the global screenshot context to: 1) remove visually dissimilar false positives; 2) confirm that grouped regions possess distinct functionalities; and 3) downsample oversized clusters ($N>5$) to retain only the most representative and confusing instances.

\noindent \textbf{Stage 3: Task Construction.} Using these refined similarity groups, we construct two challenge types:
a) \textbf{Functionality-Oriented Grounding.} We generate questions requiring the agent to localize a target region by outputting its bounding box based on an functionality requirement. We randomly select a target region from each similarity group and generate a question. For example, ``Which region should you interact to switch workspaces?'', which strictly avoids appearance-based cues.
b) \textbf{Functionality-Oriented Captioning.} We formulate multi-choice questions predicting interaction outcome. For example, ``What happens if I interact with the highlighted vertical bar?'' Options include the ground truth, \textit{hard negatives} (outcomes of visually similar regions in the same similarity group), and \textit{easy negatives} (random screen functions). Answer positions are randomized to prevent bias.

\subsection{Element-Level Functionality Understanding}
\label{sec:element task gen}

The element-level pipeline (Fig.~\ref{fig: task gen}) mirrors the region-level methodology but adapts the element discovery phase to handle fine-grained GUI components.

\noindent \textbf{Detection and Grouping.} Unlike the recursive VLM division used for regions, we employ \textbf{OmniParser-v2}~\cite{OmniParser} to detect all interactive elements. To identify visually similar candidates, we compute a cosine similarity matrix on visual embeddings extracted via \textbf{DINO-v3}~\cite{dinov3}. To remove false positives, we retain pairs with high textual similarity ($s_{text} > th_{text}$) using fuzzy matching\footnote{https://github.com/rapidfuzz/RapidFuzz} and consolidate clusters via a Disjoint Set Union (DSU) structure~\cite{DSU}..

\noindent \textbf{Refinement and Task Construction.} The candidate groups undergo the same VLM-based verification and annotation process described in Sec.~\ref{sec:region task gen} to ensure functional distinctness. Finally, we generate grounding and captioning tasks using the identical formulation strategies defined in Sec.~\ref{sec:region task gen}. These questions are tailored to element-specific interactions (e.g., ``double-click'', ``long-press''), challenging the model to distinguish between the target and ``hard negative'' elements that look similar but behave differently.

\subsection{Dataset Statistics}

\begin{table}[]
\centering
\caption{Statistics of AutoGUI-v2 evaluation tasks.}
\vspace{-1em}
\label{tab:task stats}
\resizebox{\columnwidth}{!}{%
\begin{tabular}{@{}cccc@{}}
\toprule
Task Type                & \#Tasks & \#Avg. Words of Questions & Top-3 Res.                                                                                        \\ \midrule

\rowcolor[HTML]{EFEFEF} 
Region-Level Grounding   &  442    &       20.2                &    1920x1080: 50.9\%, 1280x720: 22.2\%, 3840x2160: 5.9\%      \\

Region-Level Captioning  &  447    &       60.0                &   1920x1080: 50.1\%, 1280x720: 21.7\%, 3840x2160: 5.8\%  \\

\rowcolor[HTML]{EFEFEF} 
Element-Level Grounding  & 1076     & 24.3                      & 1920x1080: 28.3\%, 3840x2160: 22.8\%, 2560x1440: 22.8\% \\

Element-Level Captioning & 788     & 96.0                      & 1920x1080: 33.4\%, 2560x1440: 21.7\%, 3840x2160: 20.3\% \\ \bottomrule
\end{tabular}%
}
\end{table}
Our pipeline yields 3,710 hierarchical functional regions, enabling 2,753 evaluation tasks (889 region-level, 1,864 element-level). As shown in Tab.~\ref{tab:bmk comparison} and \ref{tab:task stats}, the benchmark spans multiple platforms and resolutions, testing diverse, rich functional semantics (Fig.~\ref{fig:teaser}). Further statistical details and cost analysis are provided in Appendix Sec.~\ref{sec:supp:data stats}. Task samples can be viewed in the \textit{supplementary materials}.
\section{Experiments}
\label{sec:experiments}
With AutoGUIv2, we rigorously evaluate whether existing VLMs can accurately understand the contextual functionality of GUI regions and elements.

\begin{table*}[]
\centering

\caption{\textbf{Region grounding performance and decomposition by region type and NID score.} Open-source models (GLM-4.5V, Qwen3-VL) dominate commercial leaders. The benchmark's challenge is confirmed by a universal performance drop when grounding by functionality instead of description. We also note a clear model preference for common regions types (e.g., Global Navigation) and a split NID preference (commercial models favor `Dense' context, open-source favor `Sparse'). Moreover, the lower IoU indicates that accurately outputting region boxes is significantly harder. (UI-tars and UGround do not possess a box prediction ability.}
\label{tab:exp:regiongnd}
\resizebox{\textwidth}{!}{%
\sisetup{table-format=2.1}
\begin{tabular}{@{} c *{12}{S} @{}}
\toprule
\multirow{2}{*}{\textbf{Model}} & {\multirow{2}{*}{\shortstack{\textbf{Func. Gnd.} \\ \textbf{IoU.}}}} & {\multirow{2}{*}{\shortstack{\textbf{Desc. Gnd.} \\ \textbf{Acc.}}}} & {\multirow{2}{*}{\shortstack{\textbf{Func. Gnd.} \\ \textbf{Acc.}}}} & \multicolumn{6}{c}{\textbf{Func. Gnd. Acc. by Region Type}}                                                    & \multicolumn{3}{c}{\textbf{Func. Gnd. Acc. by NID}} \\ 
\cmidrule(lr){5-10} \cmidrule(lr){11-13}
                                &                                           &                                           &                                  & {Primary Containers} & {Global Navigation} & {Content Display} & {Interaction Controls} & {Contextual Overlays} & {Others} & {Sparse}           & {Medium}          & {Dense}          \\ 
\midrule                                
Gemini-2.5-Pro-Thinking~\cite{gemini}         & \underline{20.5}                             & 77.8                                      & 77.6                                      & \underline{88.5}               & 79.6              & 74.3            & \textbf{84.0}                 & 71.0                & 62.2   & 77.5             & 76.0            & 79.7           \\
O3~\cite{o3}                              & 17.4                             & 49.2                                      & 36.7                                      & 53.8               & 34.3              & 34.3            & 44.0                 & 61.3                & 22.2   & 17.0             & 33.5            & 61.7           \\
GPT-5~\cite{gpt5}                           & 12.8                             & 36.7                                      & 31.4                                      & 38.5               & 32.5              & 34.3            & 28.0                 & 32.3                & 20.0   & 22.5             & 31.1            & 41.4           \\
Claude-Sonnet-4.5~\cite{claude}               & 12.7                             & 50.7                                      & 48.6                                      & 53.8               & 50.4              & 48.6            & 52.0                 & 51.6                & 31.1   & 50.0             & 49.1            & 46.6           \\ 
\midrule
Qwen3-VL-32B-Instruct~\cite{qwen3vl}           & \textbf{23.3}                             & \underline{92.3}                                      & \underline{84.4}                                      & \textbf{92.3}               & \underline{84.6}              & \textbf{85.7}            & \textbf{84.0}                 & \textbf{90.3}                & \textbf{73.3}   & \underline{82.4}             & \textbf{85.6}            & \textbf{85.0}           \\
Qwen3-VL-8B-Instruct~\cite{qwen3vl}            & 15.4                             & 87.3                                      & 70.1                                      & \underline{88.5}               & 69.6              & 68.6            & 76.0                 & 77.4                & 55.6   & 62.7             & 69.5            & 78.9           \\
Step-3~\cite{step3}                          & 11.0                             & 47.5                                      & 36.7                                      & 34.6               & 40.0              & 28.6            & 44.0                 & 29.0                & 24.4   & 53.5             & 29.3            & 27.8           \\
GLM-4.5V~\cite{glm45}                         & 16.0                             & \textbf{93.7}                                      & \textbf{84.6}                                      & \textbf{92.3}               & \textbf{87.1}              & \underline{77.1}            & \underline{80.0}                 & \textbf{90.3}                & \underline{68.9}   & \textbf{87.3}             & \underline{85.0}            & \underline{81.2}           \\ \midrule

UI-Tars-1.5~\cite{uitars}                      &             {--}                     & 78.3                                      & 75.8                                      & 84.6               & 76.8              & 68.6            & 72.0                 & \textbf{90.3}                & 62.2   & 74.6             & 77.8            & 74.4           \\
UGround-V1-7B~\cite{uground}                   &                {--}                  & 88.5                                      & 64.7                                      & 50.0               & 66.1              & 60.0            & 72.0                 & \underline{83.9}                & 51.1   & 63.4             & 62.3            & 69.2           \\
OS-Atlas-Base-7B~\cite{osatlas}                & 12.4                             & 72.4                                      & 56.3                                      & 61.5               & 52.9              & 62.9            & 68.0                 & \underline{83.9}                & 44.4   & 47.2             & 50.9            & 72.9           \\ 
\bottomrule
\end{tabular}
}
\end{table*}
\begin{table*}[]
\centering

\caption{\textbf{Region captioning multi-choice question accuracy.} In contrast to grounding, top-tier commercial VLMs (e.g., GPT-5, O3) dominate functionality captioning tasks, surpassing open-source models. Higher error rates on `Hard' negatives confirm models are `tricked' by context, not failing randomly.}
\label{tab:exp:regioncap}

\resizebox{\textwidth}{!}{%

\sisetup{table-format=2.1}
\begin{tabular}{@{} c *{9}{S} @{}}
\toprule

\multirow{2}{*}{\textbf{Model}} & {\multirow{2}{*}{\shortstack{\textbf{Outcome Pred.} \\ \textbf{Acc.}}}} & \multicolumn{6}{c}{\textbf{Acc. by Region Type}} & \multicolumn{2}{c}{\textbf{Error Rate by Difficulty}} \\

\cmidrule(lr){3-8} \cmidrule(lr){9-10}

                     &      & {Primary Containers} & {Global Navigation} & {Content Display} & {Interaction Controls} & {Contextual Overlays} & {Others} & {Easy} & {Hard} \\
\midrule
Gemini-2.5-Pro-Thinking~\cite{gemini} & 86.2 & 79.7 & 87.0 & \textbf{84.5} & 87.8 & \underline{90.6} & \underline{83.5} & 4.0 & 9.8 \\
O3~\cite{o3}                      & 84.0 & 90.3 & 84.9 & 74.0 & 88.5 & 85.9 & 78.7 & 5.9 & 10.1 \\
GPT-5~\cite{gpt5}                   & \textbf{88.1} & \textbf{96.6} & \textbf{88.6} & \underline{80.8} & \underline{94.0} & 90.7 & 81.2 & 2.8 & 9.1 \\
Claude-Sonnet-4.5~\cite{claude}       & 84.0 & 89.5 & 84.5 & 71.6 & \textbf{96.0} & 85.7 & 79.8 & 5.4 & 10.6 \\ \midrule
Qwen3-VL-32B-Instruct~\cite{qwen3vl}   & 80.1 & 82.8 & 79.9 & 70.3 & 90.0 & 75.0 & \textbf{85.1} & 8.2 & 11.7 \\
Qwen3-VL-8B-Instruct~\cite{qwen3vl}    & 66.3 & 70.5 & 61.7 & 77.1 & 77.8 & 80.4 & 69.6 & 11.9 & 21.8 \\

GLM-4.5V~\cite{glm45}                 & \underline{86.8} & \underline{94.5} & \underline{87.7} & 80.5 & 89.8 & \textbf{92.5} & 77.3 & 4.6 & 8.6 \\
\bottomrule
\end{tabular}

}
\end{table*}

\begin{table}[]
\centering
\caption{\textbf{Element grounding accuracy for the three referring expression types provided by  AutoGUI-v2.} Functionality-oriented grounding presents a greater challenge, harder than existing description/intent grounding benchmarks. }
\label{tab:exp:elemgnd}
\resizebox{\columnwidth}{!}{%
\begin{tabular}{@{}cccccc@{}}
\toprule
\multirow{2}{*}{Model} & \multicolumn{3}{c}{AutoGUI-v2 (ours)} & \multirow{2}{*}{\begin{tabular}[c]{@{}c@{}}ScreenSpot-v2\\ (Appearance)\end{tabular}} & \multirow{2}{*}{\begin{tabular}[c]{@{}c@{}}OSWorld-G\\ (Intent)\end{tabular}} \\ \cmidrule(lr){2-4}
& Functionality & Appearance & Intent & & \\ \midrule
Gemini-2.5-Pro-Thinking~\cite{gemini} & 67.7 & 55.5 & 57.0 & 90.0 & 39.0 \\ 
O3~\cite{o3} & 4.6 & 5.5 & 6.3 & 30.5 & 9.6 \\  
GPT-5~\cite{gpt5} & 3.8 & 4.8 & 6.6 & 26.5 & 7.8 \\ 
Claude-Sonnet-4.5~\cite{claude} & 6.6 & 5.1 & 6.4 & 16.8 & 24.7 \\ 
\midrule
Qwen3-VL-32B-Instruct~\cite{qwen3vl} & \textbf{71.1} & \textbf{80.7} & \textbf{81.9} & 95.7 & 65.1 \\ 
Qwen3-VL-8B-Instruct~\cite{qwen3vl} & 57.1 & 69.5 & 74.1 & 94.8 & 58.2 \\ 
Qwen2-VL-7B-Instruct~\cite{qwen2vl} & 12.4 & 15.0 & 17.1 & 4.5 & 58.2 \\  
Hcompany/Holo2-8B~\cite{hai2025holo2modelfamily} & 32.0 & 36.3 & 37.9 & 59.2 & 70.1 \\  
GLM-4.5V~\cite{glm45} & 67.4 & 75.4 & 74.5 & 91.8 & 50.0 \\ 
Step-3~\cite{step3} & 5.4 & 5.7 & 5.4 & 15.8 & 10.4 \\ 
\midrule
OpenCUA-72B~\cite{wang2026opencua} & \underline{67.9} & \underline{77.1} & \underline{76.8} & 92.9 & 59.2 \\ 
OpenCUA-32B~\cite{wang2026opencua} & 55.4 & 67.9 & 68.6 & 93.4 & 59.6 \\ 
OpenCUA-7B~\cite{wang2026opencua} & 52.2 & 72.1 & 71.7 & 92.3 & 55.3 \\ 
UI-Venus-Ground-72B~\cite{uivenus}  & 63.4 & 70.4 & 70.9 & 95.3 & 69.8 \\ 
UI-Venus-Ground-7B~\cite{uivenus}  & 42.1 & 54.6 & 58.4 & 94.1 & 58.8 \\ 
GUI-R1-7B~\cite{guir1}  & 54.1 & 68.9 & 70.0 & 88.1 & 42.7 \\ 
InfiGUI-G1-7B~\cite{infiguig1} & 56.6 & 69.4 & 71.1 & 93.5 & 48.0  \\ 
UI-Tars-1.5~\cite{uitars} & 54.1 & 68.9 & 70.0 & 94.2 & 56.8 \\ 
UGround-V1-7B~\cite{uground} & 14.2 & 18.9 & 22.3 & 76.5 & 42.4 \\ 
OS-Atlas-Base-7B~\cite{osatlas} & 20.8 & 28.7 & 31.3 & 85.1 & 27.7 \\ 
\bottomrule
\end{tabular}%
}
\end{table}

\begin{table*}[]
\centering
\caption{\textbf{Element grounding accuracy (\%) decomposed by action type and the proposed NID score.} Accuracy remains low for complex actions (e.g., Right-Click and Drag). Moreover, performance peaks at `Dense' density (NID), suggesting that such density may provide sufficient local context to disambiguate functionality.}
\label{tab:exp:elemgnd_decomp_exp}
\resizebox{\textwidth}{!}{%

\sisetup{table-format=2.1}

\begin{tabular}{@{} c *{11}{S} @{}}
\toprule

\multirow{2}{*}{\textbf{Model}} & {\multirow{2}{*}{\shortstack{\textbf{Func. Gnd.} \\ \textbf{Acc.}}}} & \multicolumn{7}{c}{\textbf{Acc. by Action Type}} & \multicolumn{3}{c}{\textbf{Acc. by Density Indicator NID}} \\

\cmidrule(lr){3-9} \cmidrule(lr){10-12}

& & {Left-Click} & {Hover} & {Drag} & {Right-Click} & {Double-Click} & {Long-Press} & {Type} & {Sparse} & {Medium} & {Dense} \\ 

\midrule

Gemini-2.5-Pro-Thinking~\cite{gemini} & 67.7 & \underline{68.7} & 64.6 & \underline{50.0} & 61.0 & \textbf{95.6} & \textbf{93.8} & \textbf{78.6} & 64.8 & 64.5 & 73.8  \\ 
O3~\cite{o3}                         & 4.6  & 5.4  & 2.5  & 8.3 & 0.0  & 8.7 & 18.8 & 0.0  & 2.8  & 3.8  & 7.3 \\ 
GPT-5~\cite{gpt5}                    & 3.8  & 3.6  & 2.5  & 2.8 & 0.0  & 0.0 & 56.2  & 0.0  & 2.0  & 4.9  & 4.5  \\ 
Claude-Sonnet-4.5~\cite{claude}      & 6.6  & 7.4 & 2.0 & 8.3 & 0.0 & 13.0 & 81.2  & 0.0  & 6.2 & 5.7 & 7.9  \\ 
\midrule
Qwen3-VL-32B-Instruct~\cite{qwen3vl} & \textbf{71.1} & \textbf{71.5} & \textbf{71.8} & 34.5 & 40.0 & 78.3 & \underline{87.5} & \textbf{78.6} & 64.9 & 69.3 & 79.2 \\ 
Qwen3-VL-8B-Instruct~\cite{qwen3vl}  & 57.1 & 55.9 & 57.9 & 36.1 & 61.9 & \underline{87.0}  & 81.3 & 57.1 & 49.6 & 57.9 & 63.7 \\ 
Qwen2-VL-7B-Instruct~\cite{qwen2vl}  & 12.2 & 11.2 & 8.5 & 5.7 & 22.2 & 40.0  & 77.8 & 33.3 & 6.5 & 12.1 & 18.8 \\ 

Hcompany/Holo2-8B~\cite{hai2025holo2modelfamily} & 32.6 & 32.0 & 31.6 & 19.4 & 40.0 & 54.5  & 60.0 & 42.9 & 16.5 & 40.1 & 41.0 \\ 

Step-3~\cite{step3}                  & 5.4  & 5.6  & 4.5  & 2.8  & 1.5  & 4.3 & 15.9  & 14.3 & 3.1 & 6.0 & 7.1  \\  
GLM-4.5V~\cite{glm45}                 & 24.4 & 34.0 & 27.5 & 15.3 & 17.0 & 13.4 & 35.7 & 32.4 & 21.8 & 36.0 & 24.8 \\  
\midrule

OpenCUA-72B~\cite{wang2026opencua} & \underline{67.9} & 67.8 & \underline{68.7} & 47.1 & \textbf{68.4} & 76.2 & 86.7 & 69.2 & 64.2 & 64.2 & 75.4 \\ 
OpenCUA-32B~\cite{wang2026opencua} & 55.4 & 55.6 & 57.0 & 38.9 & 52.4 & 73.9 & 31.2 & 50.0 & 48.7 & 51.1 & 66.4 \\  
OpenCUA-7B~\cite{wang2026opencua} & 52.2 & 50.2 & 54.5 & 38.9 & 57.1 & 69.6 & 81.3 & 50.0 & 44.5 & 49.7 & 62.5 \\ 
UI-Venus-Ground-72B~\cite{uivenus}  & 63.4 & 65.1 & 60.4 & 41.7 & \underline{66.7} & 73.9 & \underline{87.5} & \underline{71.4} & 65.4 & 57.9 & 67.0 \\ 
UI-Venus-Ground-7B~\cite{uivenus}  & 42.1 & 42.1 & 40.8 & 30.6 & 28.6 & 60.9 & 81.3 & 50.0 & 38.0 & 41.1 & 47.3 \\ 
GUI-R1-7B~\cite{guir1}  & 45.4 & 46.8 & 41.0 & 36.1 & 38.1 & 73.9 & \underline{87.5} & 35.7 & 37.6 & 45.6 & 53.0 \\ 
InfiGUI-G1-7B~\cite{infiguig1} & 56.6 & 55.7 & 56.7 & \textbf{55.6} & 38.1 & 78.3 & 81.3 & 57.1 & 51.5 & 54.6 & 63.7 \\ 
UI-Tars-1.5~\cite{uitars}             & 54.1 & 55.8 & 52.5 & 44.4 & 47.6 & 60.9 & 68.8 & 28.6 & 50.0 & 54.1 & 58.3 \\ 
UGround-V1-7B~\cite{uground}         & 14.2 & 14.4 & 10.4 & 16.7 & 9.5 & 21.7 & 68.8 & 28.6 & 12.7 & 12.6 & 17.5 \\ 
OS-Atlas-Base-7B~\cite{uground}      & 20.8 & 21.2 & 15.7  & 22.9 & 28.6 & 30.4 & 81.3 & 28.6 & 13.9  & 18.6 & 29.9 \\ 
\bottomrule
\end{tabular}%
}
\end{table*}

\begin{table*}[]
\centering
\caption{\textbf{Element-level functionality captioning multi-choice question accuracy.} Commercial VLMs lead in this interaction prediction task, revealing a divergence from models fine-tuned for grounding. Additionally, performance correlates with an action's visual feedback, succeeding on overt actions (Type) but failing on subtle ones (Hover). Critically, higher error rates on `Hard' negatives than `Easy' ones indicate a gap in context-aware functionality reasoning.}
\label{tab:exp:elemcap}

\resizebox{\textwidth}{!}{%

\sisetup{table-format=2.1}
\begin{tabular}{@{} c *{10}{S} @{}}
\toprule
\multirow{2}{*}{\textbf{Model}} & {\multirow{2}{*}{\shortstack{\textbf{Outcome Pred.} \\ \textbf{Acc.}}}} & \multicolumn{7}{c}{\textbf{Acc. by Action Type}} & \multicolumn{2}{c}{\textbf{Error Rate by Difficulty}} \\
\cmidrule(lr){3-9} \cmidrule(lr){10-11}
                     &      & {Left-Click} & {Hover} & {Drag} & {Right-Click} & {Double-Click} & {Long-Press} & {Type} & {Easy} & {Hard} \\
\midrule
Gemini-2.5-Pro-Thinking~\cite{gemini} & \textbf{70.3} & \underline{66.0} & \underline{61.5} & \underline{71.0} & \textbf{80.0} & 69.6 & \underline{76.9} & 89.3 & 11.7 & 18.0 \\
O3~\cite{o3}                      & 65.4 & 59.5 & 54.5 & 68.5 & \underline{70.9} & 62.5 & \textbf{84.6} & \underline{90.3} & 16.6 & 18.0 \\
GPT-5~\cite{gpt5}                   & 66.5 & 59.0 & 56.0 & \underline{71.0} & 65.5 & \underline{71.4} & \textbf{84.6} & \textbf{91.4} & 16.6 & 16.9 \\
Claude-Sonnet-4.5~\cite{claude}       & \underline{66.9} & 59.5 & \textbf{63.5} & \underline{71.0} & 69.1 & 57.1 & \textbf{84.6} & 86.0 & 15.4 & 17.8 \\ \midrule
Qwen3-VL-32B-Instruct~\cite{qwen3vl}   & 61.3 & 57.5 & 50.5 & \textbf{72.2} & 50.9 & 67.9 & 38.5 & 79.6 & 15.0 & 23.7 \\
Qwen3-VL-8B-Instruct~\cite{qwen3vl}    & 57.1 & 58.0 & 48.5 & 59.9 & 47.3 & 64.3 & 46.2 & 72.0 & 15.7 & 27.2 \\
Step-3~\cite{step3}                  & 52.8 & 43.5 & 47.0 & 56.8 & 65.5 & 60.7 & 38.5 & 65.6 & 21.4 &  25.8 \\
GLM-4.5V~\cite{glm45}                 & 64.8 & \textbf{67.0} & 55.0 & 69.1 & 58.2 & \textbf{73.2} & 30.8 & 76.3 &    14.6  &  20.6  \\
\bottomrule
\end{tabular}
}
\end{table*}
\begin{figure*}[th]
  \centering
   \includegraphics[width=1.0\linewidth]{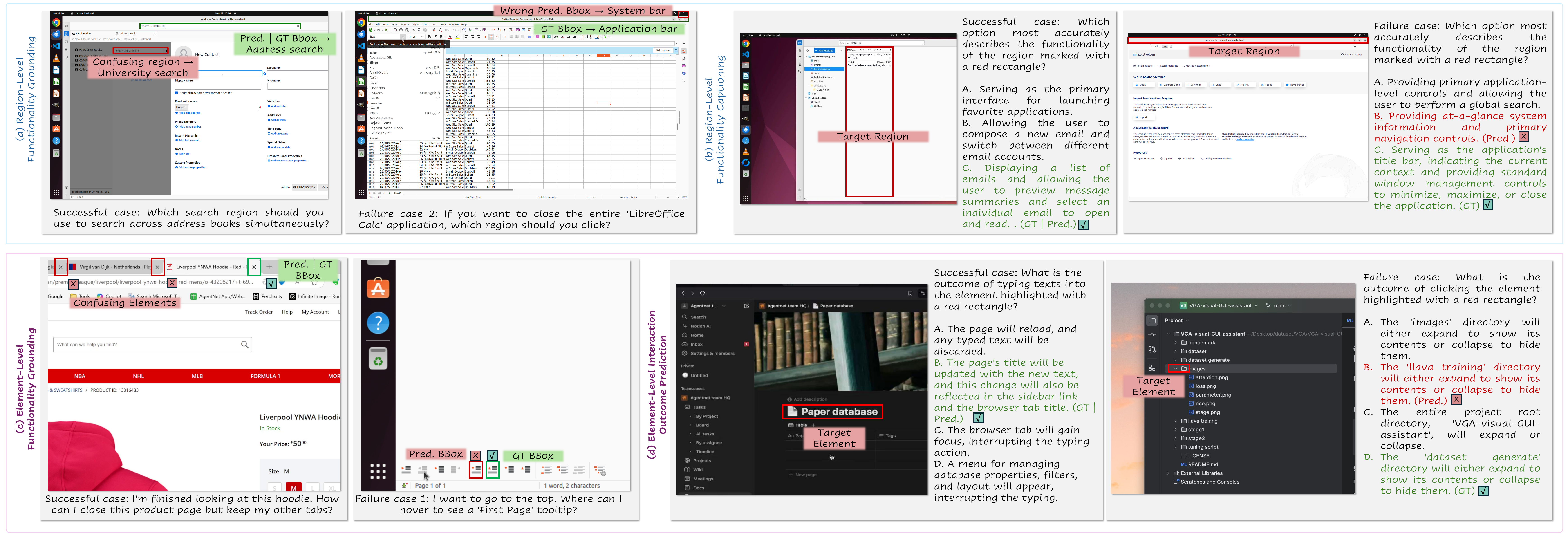}

   \caption{\textbf{Functionality understanding cases of Gemini-2.5-Pro-Thinking.} This VLM successfully understand common GUI targets (e.g., search region) but fails on complex or stateful ones (e.g., long software status bar).}
   \label{fig:case analysis}
\end{figure*}
\subsection{Experimental Settings}
\noindent \textbf{Evaluated Models.}
We evaluate a comprehensive suite of commercial and open-source VLMs, categorized as follows: \textbf{a) Commercial Models:} Leading VLMs accessed via developer platforms, including Gemini-2.5-Pro-Thinking~\cite{gemini}, OpenAI O3~\cite{o3}, GPT-5~\cite{gpt5}, and Claude-Sonnet-4.5~\cite{claude}. \textbf{b) General-Purpose Open-Source:} Models such as Qwen3-VL~\cite{qwen3vl}, GLM-4.5V~\cite{glm45}, and Step-3~\cite{step3}, also accessed via cloud APIs. \textbf{c) GUI-Grounding-Oriented Open-Source:} Models including UI-Tars-1.5~\cite{uitars} (based on Qwen2.5-VL~\cite{qwen2vl}), UGround~\cite{uground}, and OS-Atlas~\cite{osatlas} (both based on Qwen2-VL~\cite{qwen2vl}). These are deployed on an A100 GPU via HuggingFace Inference Endpoints\footnote{https://endpoints.huggingface.co/}. All models are evaluated at their native resolutions using performance-tuned prompts (see Appendix Sec.~\ref{sec:supp:eval details}).

\noindent \textbf{Evaluation Metrics.} Grounding is measured by IoU and Grounding Center Accuracy (Acc.)~\cite{cheng2024seeclick,li2025autoguiscalingguigrounding}, where $\text { Acc }=  \frac{1}{N} \sum_{i=1}^N \mathbf{1}\left(\text {pred}_i \text { inside GT } \text {bbox}_i\right) \times 100 $ where $\mathbf{1}$ is an indicator function and $N$ the number of test samples. This metric represents the percentage of samples with the predicted points falling in the ground truth bounding boxes. For captioning tasks, we report standard multi-choice question answering accuracy~\cite{MMMU, MathVista}.

To analyze performance under varying visual complexity, we decompose metrics by region/action type and propose the \textbf{Normalized Interference Density (NID)}. NID quantifies the visual clutter surrounding a target $e_{\text{target}}$ with bounding box $B_{\text{target}}$ (width $W$, height $H$). We first define an ``analysis region'' $R_{\text{analysis}}$, centered on the target, with a scaled width $W_R=W\cdot(1+2\alpha)$ and height $H_R=H\cdot(1+2\alpha)$, where $\alpha=1.0$ is a fixed expansion factor. NID is the count of other element centers $e_j$ within this analysis region: 
$$ 
NID = \sum_{e_j \in E \setminus \{e_{\text{target}}\}} \mathbf{1}\left(\text{center}(e_j) \in R_{\text{analysis}}\right).
$$ 
We use NID tertiles to objectively classify samples into \textbf{Sparse}, \textbf{Medium}, and \textbf{Dense} groups, enabling a fine-grained analysis of model robustness to visual interference.

\subsection{Benchmarking Results}

\subsubsection{Region-Level Functionality Understanding}
To evaluate the models' capability in parsing macro-scale GUI structures and functional areas, we conduct assessments on region-level grounding and captioning tasks. The results are summarized in Tab.~\ref{tab:exp:regiongnd} and Tab.~\ref{tab:exp:regioncap}.

\noindent \textbf{Open-source VLMs specialized in GUIs dominate region grounding.}
Specialized open-source models demonstrate a significant advantage in localizing functional regions. As shown in Tab.~\ref{tab:exp:regiongnd}, Qwen3-VL-32B-Instruct~\cite{qwen3vl} and GLM-4.5V~\cite{glm45} achieve remarkable functionality grounding accuracies of 84.4\% and 84.6\%, respectively. In stark contrast, general-purpose commercial models struggle significantly in this metric; for instance, GPT-5~\cite{gpt5} only attains 31.4\%, and even the robust Gemini-2.5-Pro-Thinking~\cite{gemini} (77.6\%) trails the leading open-source models. This performance gap suggests that while commercial models possess strong general capabilities, they may lack the specific ability to pinpoint large-scale GUI regions, whereas large-scale fine-tuning with GUI data appears advantageous.

\noindent \textbf{Functionality-oriented grounding presents a greater challenge than description-oriented grounding.}
Comparing the ``Desc. Gnd. Acc.'' and ``Func. Gnd. Acc.'' columns in Tab.~\ref{tab:exp:regiongnd} reveals a consistent performance drop across almost all models when shifting from visual descriptions to functionality. For example, UGround-V1-7B~\cite{uground} drops from 88.5\% to 64.7\%, and OS-Atlas-Base-7B~\cite{uground} drops from 72.4\% to 56.3\%. This trend indicates that mapping explicit visual attributes (e.g., ``the blue sidebar'') to coordinates is inherently less challenging than deducing regions from abstract functional descriptions, which requires a deeper level of semantic processing beyond mere visual matching.

\noindent \textbf{Standardized regions are easier to localize than Others regions.}
Breaking down performance by the region types exemplified in Fig.~\ref{fig:teaser}, Tab.~\ref{tab:exp:regiongnd} shows that the models consistently achieve high accuracy on standardized regions such as Primary Containers and Interaction Controls. However, regarding the Others category (e.g., color pickers and logos)—which typically comprises irregular layout regions—we observe model performance generally drops by at least 5\% compared with the best-performing region types. Interestingly, Claude-Sonnet-4.5 ~\cite{claude} exhibits strong performance on Interaction Controls, suggesting that while it may struggle with broader regions, it retains strong capabilities in recognizing interactive widgets, such as filters, search regions, and paginations.

\noindent \textbf{Density preference varies by model type.}
The NID analysis in Tab.~\ref{tab:exp:regiongnd} reveals an interesting divergence. Commercial models (e.g., O3~\cite{o3}, GPT-5~\cite{gpt5}) and GUI-specialized models (e.g., OS-Atlas-Base-7B~\cite{osatlas}) generally exhibit their highest accuracy in Dense environments (e.g., O3~\cite{o3} achieves 61.7\% in Dense vs. 17.0\% in Sparse). In contrast, general-purpose open-source models (e.g., Step-3~\cite{step3}, GLM-4.5V~\cite{glm45}) tend to perform best in Sparse scenarios. This phenomenon probably suggests that commercial and domain-specific models might benefit from the richer contextual cues provided by neighboring layout structures in dense settings, whereas general open-source models seem to prefer cleaner visual inputs with less distractor interference. For a more detailed analysis of this trend, please refer to the Appendix Sec.~\ref{sec:supp:eval details}.

\noindent \textbf{Hard negatives reveal robustness gaps in functional reasoning.}
To verify the rigor of our evaluation, we analyze the Error Rate by Difficulty in Tab.~\ref{tab:exp:regioncap}. Across all models, the error rate on Hard samples (distractors from the same similarity group) is consistently higher than on Easy samples. For instance, GPT-5~\cite{gpt5} has a Hard error rate of 9.1\% versus an Easy error rate of 2.8\%. This persistent gap indicates that models are not failing randomly; they are more frequent to be ``tricked'' by functionally plausible distractors, highlighting that functional discrimination remains a non-trivial challenge even at the region level.

\subsubsection{Element-Level Functionality Understanding}

The VLMs are evaluated on our element-level tasks to assess fine-grained functional understanding. The results are presented in Tables~\ref{tab:exp:elemgnd},~\ref{tab:exp:elemgnd_decomp_exp}, and~\ref{tab:exp:elemcap}.

\noindent \textbf{Open-source VLMs lead in grounding accuracy.}
Tab.~\ref{tab:exp:elemgnd} shows that the open-source Qwen3-VL-32B-Instruct~\cite{qwen3vl} emerges as the top performer, surpassing leading commercial models like Gemini-2.5-Pro-Thinking~\cite{gemini}.
We attribute this gap to Qwen3-VL's specialized fine-tuning on GUI agent tasks.
Similarly, GUI-oriented VLMs (e.g., OpenCUA) outperform general commercial models, suggesting that large-scale, domain-specific fine-tuning enhances functional understanding, though it remains insufficient to master our benchmark.

\noindent \textbf{Functionality-based grounding is significantly harder than appearance or intent-based grounding.} To quantify the benchmark's difficulty, we test the VLMs on parallel tasks using appearance or action-intent prompts for the same elements. As shown in Tab.~\ref{tab:exp:elemgnd}, performance on functionality-based tasks is consistently lower across all models. This disparity, combined with higher scores on existing benchmarks (ScreenSpot-v2~\cite{osatlas}, OSWorld-G~\cite{xie2025scalingcomputerusegroundinguser}), confirms that understanding functionality is a distinct and more difficult challenge than localizing based on appearance or intent cues.

\noindent \textbf{Models struggle with complex and implicit actions.}
The action-type breakdown (Tab.~\ref{tab:exp:elemgnd_decomp_exp}) reveals a clear pattern: models perform best on actions with explicit outcomes, such as Long-Press and Type. Conversely, accuracy plummets for complex or implicit actions, particularly Right-Click (near-zero for most commercial models) and Right-Click. This suggests models fail when functionality is not immediately obvious. While Qwen3-VL and OpenCUA show the strongest overall grounding, their strengths are non-uniform, highlighting inconsistent mastery of action semantics.

\noindent \textbf{NID analysis reveals an increasing trend.}
The NID breakdown (Tab.~\ref{tab:exp:elemgnd_decomp_exp}) shows an overall increasing trend except GLM-4.5V and UI-Venus-Ground-72. For most of the models, when the number of surrounding elements increases (i.e., NID rises), the grounding accuracy for this group also increases. This suggests that element-level functionality understanding likely relies on a rich context, which means determining the functionality of an element in a functional group is easier than that of an isolated element.

\noindent \textbf{Commercial VLMs excel at captioning, revealing a divergence.} In sharp contrast to grounding results, top-tier commercial models (Gemini-2.5-Pro-Thinking) dominate the functional captioning task (Tab.~\ref{tab:exp:elemcap}).
This suggests a clear divergence: GUI-specific fine-tuning improves localization (``where is it?''), but the broad world knowledge of large commercial models is more effective for abstractly articulating purpose (``what does it do?'').

\noindent \textbf{Captioning accuracy correlates with action types.} The captioning breakdown by action type (Tab.~\ref{tab:exp:elemcap}) shows models perform best on actions with overt visual state changes (e.g., Type, Long-Press). Hover is a consistent failure point, as its subtle feedback (e.g., a tooltip) provides a weaker signal for models to learn the associated function.

\noindent \textbf{Error analysis shows models fail on plausible distractors.} Our multi-choice setup includes ``easy'' (irrelevant) and ``hard'' (visually/semantically similar) negatives. Across all models, the Hard error rate is consistently higher than the Easy rate. This demonstrates that models are not failing randomly; they are actively tricked by plausible-but-incorrect functional descriptions (Fig.~\ref{fig:case analysis}), highlighting a critical gap in nuanced, context-aware reasoning.

\subsubsection{Case Visualization}
Qualitative analysis of Gemini-2.5-Pro-Thinking (Fig.~\ref{fig:case analysis}) reveals its capabilities and limitations. For grounding, the VLM excels at common, unambiguous elements like search bars and `close' buttons. However, it falters on components with complex or abstract functions, such as long status bars or the tiny, specialized icons in productivity software.

For functionality captioning, the model accurately identifies outcomes for navigational elements (e.g., an email preview list) but struggles with dynamic, stateful controls. For instance, it fails to predict the behavior of application control bars or the `chevron' icon for a collapsible folder. These cases, with more visualized in the Appendix, indicate that while VLMs grasp common functions, they currently lack a robust, holistic understanding of complex GUI dynamics.
\section{Conclusion}
\label{sec:conclusion}
We propose AutoGUI-v2, a comprehensive benchmark for evaluating context-aware GUI functionality understanding at both the region and element levels. We benchmark various leading VLMs and provide insightful analysis. Our results reveal a key ``grounding-reasoning divergence" and show that a deep understanding of complex interactions and GUI dynamics remains highly challenging for current models. We hope our work will inspire future efforts to improve GUI agents. For social impacts and limitations of our work, please refer to Sec.~\ref{sec:supp:social impact} and Sec.~\ref{sec:supp:limitations}.

{
    \small
    \bibliographystyle{ieeenat_fullname}
    \bibliography{main}
}

\startcontents

\clearpage
\setcounter{page}{1}
\maketitlesupplementary

\section{Implementation Details of AutoGUI-v2 Datasets}

\subsection{GUI Data Sources}
\label{sec:supp:data sources}
The following GUI data sources are utilized by AutoGUI-v2 to generation functional regions along with functionality descriptions:

\noindent \textbf{AMEX}, also named Android Multi-annotation EXpo~\cite{amex}, is a comprehensive, large-scale dataset designed for mobile GUI agents, providing 104K high-resolution screenshots from 110 popular mobile applications. As element-level functionality understanding task generation requires detecting similarity groups (Sec.~\ref{sec:element task gen}), not all screenshots are qualified as a task source. Therefore, we need to perform filtering before task generation. Specifically, we first utilize the data pre-processing pipeline in UIpro~\cite{UIPro} to obtain the screenshots of the test set of this benchmark, then employ OmniParser-v2~\cite{OmniParser} to detect all elements for the screenshots, and finally find all similarity groups. Subsequently, 120 are randomly sampled from the screenshots with at least one similarity group to undergo the region division procedure in sec.~\ref{sec:method}. This approach can ensure that the source screenshots contain rich content for both region and element-level task generation.

\noindent \textbf{AndroidControl}~\cite{androidcontrol} possesses 14,548 unique tasks over 833 Android apps, allowing researchers to conduct comprehensive analysis of agent performance. Likewise, the data pre-processing pipeline provided by UIPro~\cite{UIPro} is first used to obtain test set screenshots and filtering through similarity group calculation is subsequently performed to select 120 qualified screenshots for task generation.

\noindent \textbf{ScreenSpot-Pro}~\cite{screenspotpro} is a benchmark designed to rigorously evaluate the grounding capabilities of VLMs in high-resolution professional settings. This benchmark comprises high-resolution GUI screenshots from 23 applications across five industries and three operating systems (i.e., MacOS, Windows, and Linux). As the screenshots of this benchmark already contain super-rich hierarchical content, we directly detect functional regions for 271 randomly selected from the whole 1,581 screenshots.

\noindent \textbf{AgentNet}~\cite{wang2026opencua} is a large-scale computer-use task dataset spanning 3 operating systems (i.e., MacOS, Windows, and Ubuntu) and 200+ applications and websites. We randomly select 120 screenshots from this benchmark for functional region division.

\noindent \textbf{MMBenchGUI}~\cite{mmbenchgui} is a hierarchical benchmark for evaluating GUI automation agents across Windows, MacOS, Linux, iOS, Android, and Web platforms. This benchmark consists of four task types—GUI Content Understanding, Element Grounding, Task Automation, and Task Collaboration. As this data source is significantly diverse, all 1,856 screenshots are used for functional region division and task generation.

\noindent \textbf{OSWorld-G}\footnote{https://huggingface.co/datasets/MMInstruction/OSWorld-G}~\cite{osworldg} is a comprehensive benchmark comprising 564 annotated GUI element grounding tasks across diverse task types including text matching, element recognition, layout understanding, and precise manipulation. The 250 unique screenshots are used for region division and task generation.

\subsection{Functional Region Division Details}
\label{sec:supp:division details}

This sub-section provides a detailed technical exposition of our automated hierarchical functional region annotation pipeline, as implemented in our Python script. The primary challenge, as noted in the main paper, is that VLM outputs, while semantically rich, often lack the geometric precision required for a rigorous benchmark. A VLM may generate bounding boxes that are shifted, incomplete, or excessively large, failing to perfectly encompass all necessary GUI elements.

Our methodology is explicitly engineered to mitigate this imprecision through a multi-stage, recursive ``propose-verify-refine'' loop. The entire process is encapsulated within the \texttt{FunctionalRegionAnnotator} class, which orchestrates the hierarchical decomposition of a GUI screenshot.

\subsubsection{Algorithm: Recursive ``Divide-and-Verify''}
The core of our pipeline is the \texttt{annotate\_image} function, which implements a depth-first traversal algorithm. The process begins by placing the entire screenshot (the root node, e.g., \texttt{'0-0'}) onto a processing stack. The annotator then iteratively processes nodes from this stack.

\begin{itemize}
    \item \textbf{Initialization:} A node $n$ is popped from the stack, representing a region $\mathbf{I}_{region}$ (initially, the entire GUI).
    \item \textbf{Proposal Generation:} The region $\mathbf{I}_{region}$ is sent to our primary annotation VLM (\texttt{Gemini-2.5-Pro-Thinking}) with the annotation prompt $p_{anno}$ (\texttt{ANNO\_PROMPT\_V2\_EN}, shown in Tab.~\ref{tab:division prompt}). The VLM is prompted to return a list of $K$ child regions, $\{r_i\}_{i=1}^K$, each with a bounding box $B_i$, functionality $F_i$, description $D_i$, and a divisibility flag $d_i$.
    \item \textbf{Iterative Refinement Loop:} This is the most critical stage. We do not naively accept the VLM's first proposal. Instead, we enter an iterative refinement loop (controlled by \texttt{max\_refine}) to find the optimal division for $\mathbf{I}_{region}$.
\end{itemize}

\subsubsection{Automated Quality Verification}
Within the refinement loop, each proposal (the set of $K$ child regions) is subjected to a rigorous verification process using a separate, high-speed checking\_model (Gemini-2.5-Pro-Thinking) and the scoring prompt $p_{scoring}$ (CHECK\_REGION\_COMPLETENESS\_PROMPT, shown in Tab.~\ref{tab:checking prompt}). The full details will be delineated in Sec.~\ref{sec:supp:scoring details}.

\subsubsection{Recursion, Termination, and Caching}
Once a "best" proposal is selected, its child regions are processed:
\begin{itemize}
    \item \textbf{Recursion:} Children $r_i$ that are marked as dividable ($d_i = \text{True}$) and are larger than a minimum size threshold (its width and height are both greater than 30 pixels) are pushed onto the processing stack for further decomposition in the next iteration.
    \item \textbf{Termination:} A branch of the hierarchy terminates when a region is marked as non-dividable, falls below the size threshold, or the maximum depth (\texttt{max\_level}) is reached.
    \item \textbf{Caching:} Every processed node (both parent and child) has its metadata, cropped image, and raw VLM responses saved to disk (\texttt{\_write\_node\_cache}). This creates a complete, auditable trace of the generation process and populates the \texttt{tree.json} and \texttt{stack.json} files, which are designed to feed the interactive web UI used for Stage 2 (Human Correction, detailed in Sec.~\ref{sec:supp:reanno details}).
\end{itemize}

This automated pipeline, therefore, serves as the critical "Stage 1" of our full methodology. It is designed not to achieve perfection, but to produce a high-quality, verified, and hierarchically structured "draft" annotation. This draft significantly reduces the high cost and cognitive load of the manual correction (Stage 2) and re-annotation (Stage 3) steps, forming the foundation of our scalable VLM-human collaborative system.

\subsection{Functional Region Scoring Details}
\label{sec:supp:scoring details}

Within the refinement loop introduced in Sec.~\ref{sec:supp:division details}, each proposal (the set of $K$ child regions) is subjected to a rigorous verification process using a separate, high-speed \texttt{checking\_model} (\texttt{Gemini-2.5-Pro-Thinking}) and the scoring prompt $p_{scoring}$ (\texttt{CHECK\_REGION\_COMPLETENESS\_PROMPT}, shown in Tab.~\ref{tab:checking prompt}).

For each proposed child $r_i$, we assess two key metrics:
\begin{enumerate}
    \item \textbf{Completeness ($s^{\text{comp}}_i$):} The checking model evaluates the cropped child region $I_i$ within the context of its parent $I_{region}$ (which is marked with a red rectangle for reference). It returns a score from 0-3, assessing if the region is fully visible and functionally coherent.
    \item \textbf{Boundedness ($s^{\text{bound}}_i$):} The model provides a binary (Yes/No) judgment on whether the bounding box $B_i$ tightly frames the functional elements without excessive padding or cropping.
\end{enumerate}

A proposal is "accepted" if the average completeness score $\frac{1}{K}\sum s^{\text{comp}}_i$ meets our quality threshold (e.g., $\ge 2.5$) AND the ratio of "bounded" children $\frac{1}{K}\sum \mathbf{1}(s^{\text{bound}}_i = \text{Yes})$ meets its threshold (e.g., $\ge 0.8$).

If a proposal fails, the system re-prompts the annotation VLM (with a slightly higher temperature) to generate a new set of $K$ regions, up to $N = \texttt{max\_refine}$ times (e.g., $N=3$). If no proposal passes the thresholds, the system selects the one with the highest combined score, ensuring the best possible automated result is preserved.

\subsection{Functional Region Re-Annotation Details}
\label{sec:supp:reanno details}

The automated annotation pipeline (Stage 1) is designed to generate a high-quality, hierarchical "draft" of the functional regions. However, We find that the VLM-generated bounding boxes lack the pixel-perfect precision required for a rigorous benchmark. Stage 2 addresses this by introducing a sophisticated, human-in-the-loop (HITL) correction workflow, enabled by a custom web application.

\subsubsection{Annotation Correction Server}
\begin{figure*}[th]
  \centering
   \includegraphics[width=1.0\linewidth]{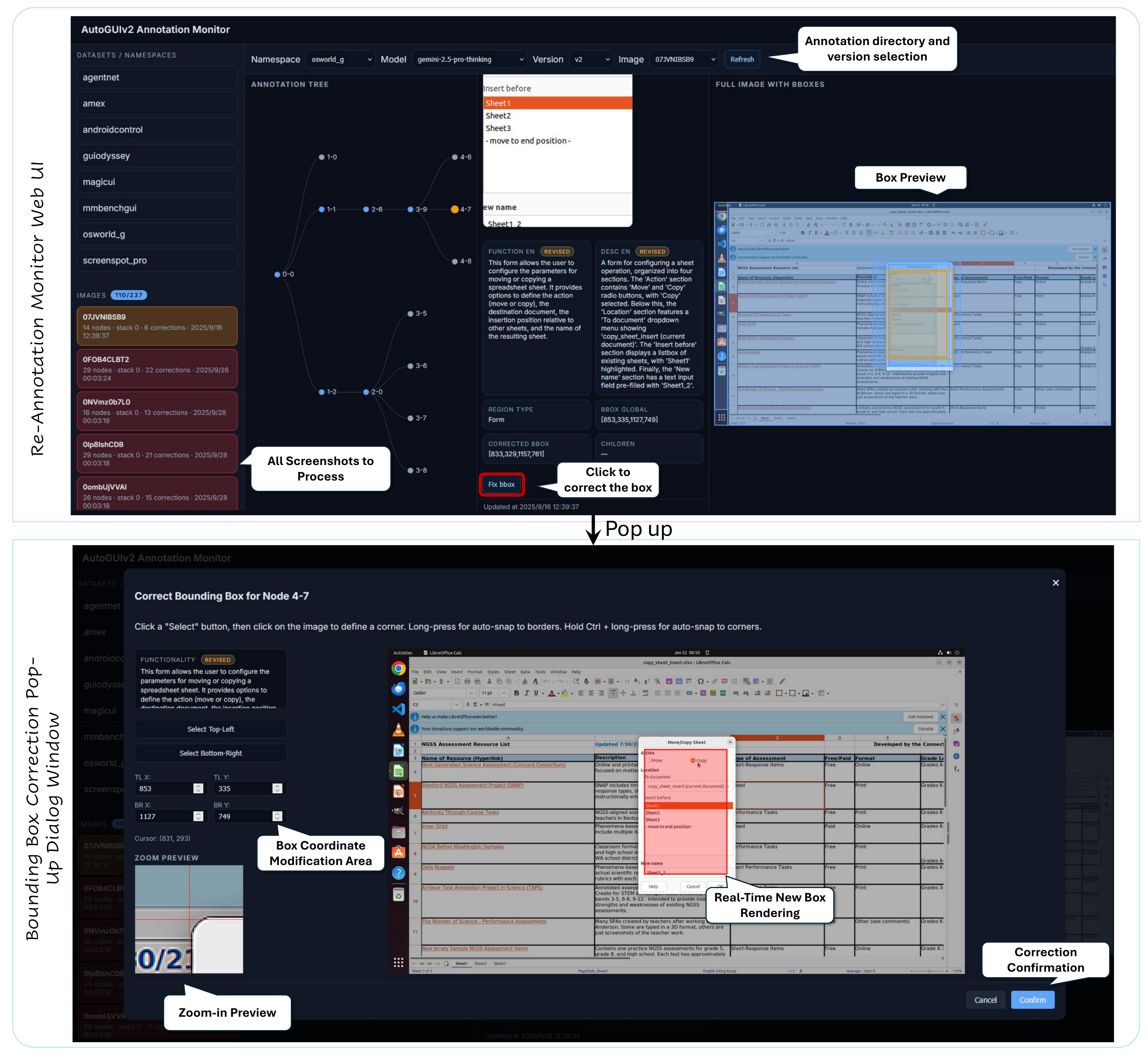}

   \caption{Illustration of the functional region bounding box correction web UI of AutoGUI-v2. After detecting functional regions in Sec.~\ref{sec:supp:division details}, the authors deploy this web UI to inspect all annotation results, and then manually correct the bounding boxes by re-selecting the top-left and bottom-right corners.}
   \label{fig:re-anno web ui}
\end{figure*}
The core of the HITL system is a \texttt{FastAPI} web server (Fig.~\ref{fig:re-anno web ui}). Its primary responsibility is to serve the interactive HTML/JavaScript frontend and provide a RESTful API to interface with the annotation data cache.

\begin{itemize}
    \item \textbf{Data Discovery:} The server intelligently discovers the annotation cache by searching a prioritized list of common paths (e.g., CLI arguments, environment variables, default paths). It parses the cache's complex directory structure (e.g., \texttt{/namespace/model/version/image\_id/}) to locate all processed images and their corresponding \texttt{tree.json} files.
    
    \item \textbf{API Endpoints:} The server exposes endpoints to securely read and write annotation data. Key endpoints include:
    \begin{itemize}
        \item \texttt{GET} \url{/api/images}: Lists all available images for correction, dynamically calculating statistics like \texttt{nodes\_count} and \texttt{corrections\_count} (by searching for \texttt{\_fix*.json} files) to populate the annotator's dashboard.
        \item \texttt{GET} \url{/api/image/.../tree}: Loads the \texttt{tree.json} file to visualize the full hierarchy.
        \item \texttt{GET} \url{/api/image/.../node/{node\_id}:} Fetches the detailed metadata for a single node, merging the original VLM output with any existing correction or re-annotation files.
        \item \texttt{POST} \url{/api/image/.../node/{node\_id}/correct}: The critical "save" endpoint. This function, \texttt{save\_corrected\_node}, receives the new, human-verified \texttt{new\_bbox} coordinates.
    \end{itemize}
    
    \item \textbf{Non-Destructive Saving:} To maintain full provenance and allow for easy comparison, the system employs a non-destructive correction process. When an annotator saves a correction, the server does not overwrite the original VLM-generated metadata. Instead, it reads the original file, updates the \texttt{bbox\_global} and \texttt{bbox\_global\_norm} fields, adds a \texttt{correction\_info} block, and saves the result as a \textit{new file} with a timestamped \texttt{\_fix} suffix (e.g., \texttt{0-0\_fix20251116.json}). The \texttt{read\_node} endpoint is designed to find and load this "fixed" version if it exists.
\end{itemize}

\subsubsection{Interactive Correction Interface}
The frontend is a single-page application (SPA) designed for maximum annotator efficiency. It features a three-panel layout:
\begin{enumerate}
    \item \textbf{Image List:} A searchable list of all GUIs, with status indicators showing the percentage of nodes that have been corrected.
    \item \textbf{Hierarchy Panel:} A D3.js-based interactive tree visualization of the selected GUI's full functional hierarchy. Clicking a node in this tree loads its metadata in the details panel.
    \item \textbf{Full Image Panel:} A view of the entire root screenshot with all VLM-proposed bounding boxes rendered on an SVG overlay.
\end{enumerate}

\subsubsection{Convenient Bounding Box Correction Workflow}
The correction process is initiated when an annotator selects a node and opens the "Fix BBox" modal. This interface is equipped with advanced tools to make correction fast and precise:

\begin{itemize}
    \item \textbf{Full-Context View:} The modal displays the \textit{full root image}, not the small region crop. This provides the annotator with complete context to make an accurate judgment about the region's true boundaries.
    
    \item \textbf{Sharpened Zoom Preview:} A real-time, magnified preview follows the annotator's cursor. This feature uses \texttt{OpenCV.js} to draw the relevant portion of the image onto a canvas and applies a 2D sharpening filter (\texttt{cv.filter2D}). This enhancement makes it significantly easier to identify precise UI element borders.
    
    \item \textbf{Auto-Snap to Edges (Long-Press):} For rapid, perfect alignment, the annotator can long-press. This triggers our "auto-snap" feature, which uses \texttt{OpenCV.js} to perform Canny edge detection (\texttt{cv.Canny}) and a probabilistic Hough line transform (\texttt{cv.HoughLinesP}) in a local window around the cursor. The system then calculates the closest point on the nearest detected line and snaps the coordinate to it.
    
    \item \textbf{Auto-Snap to Corners (Ctrl + Long-Press):} For perfect corner alignment, annotators can use \texttt{Ctrl} + long-press. This triggers an alternative snap mode that uses the "Good Features to Track" algorithm (\texttt{cv.goodFeaturesToTrack}) to find strong Harris corners in the local window, snapping the cursor to the nearest detected corner.
\end{itemize}

When the annotator confirms the new coordinates, the UI `POST`s the data to the server, which saves the non-destructive file. The UI then automatically refreshes the image card and node details, providing immediate feedback on the correction.

\subsubsection{Functionality Re-Annotation}

The human-in-the-loop correction guarantees the geometric precision of our bounding boxes. However, this necessary step creates a semantic misalignment: the original, VLM-generated descriptions ($F_i, D_i$) from Stage 1 now correspond to a different, imprecise crop. Stage 3 is the critical final pass designed to re-align the semantic annotations with the new, human-verified geometry. This is achieved by prompting \textbf{Gemini-2.5-Pro-Thinking} to revise the annotations based on the corrected bounding boxes.

The re-annotation process is orchestrated by a batch script. This script does not re-process the entire dataset, but only the nodes that have been manually corrected.

\begin{enumerate}
    \item \textbf{Correction Discovery:} The re-annotating script recursively scans the cache directory. It searches for all human-corrected metadata files, which are identified by their \texttt{\_fix*.json} or \texttt{\_meta\_fix*.json} suffix.
    \item \textbf{Task Queuing:} For each corrected file found, the script generates a \texttt{CorrectionTask} object. This object contains paths to the original root image, the node's original metadata, and the human-corrected file.
    \item \textbf{Parallel Execution:} The script processes this queue in parallel using a \texttt{multiprocessing.Pool}. Each worker executes the core \texttt{reannotate\_node} function for a given task.
\end{enumerate}

The core of this functionality re-annotation is a specialized VLM prompt (\texttt{REANNOTATION\_PROMPT\_TEMPLATE} shown in Tab.~\ref{tab:reanno prompt}) that provides the model with a rich, multi-modal context to perform an expert revision. For each node, the \texttt{reannotate\_node} function prepares the following inputs:

\begin{itemize}
    \item \textbf{The Full-Screen Context:} The original \texttt{root.png} is loaded, and the new, human-verified bounding box (from the \texttt{\_fix*.json} file) is drawn onto it as a conspicuous red rectangle.
    \item \textbf{The Corrected Crop:} The script uses the human-verified coordinates to crop the \texttt{root.png} again, producing a new, geometrically-accurate image of the functional region.
    \item \textbf{The Flawed Original Annotations:} The \texttt{previous\_functionality} and \texttt{previous\_description} are extracted from the original node metadata.
\end{itemize}

The VLM is then prompted to act as a "UI/UX analyst" and is explicitly told that a human has corrected the bounding box, implying the previous annotations may be "incorrect, either containing hallucinated details or missing important discernible details." The VLM is instructed to ignore the previous text if it conflicts with the new visual evidence and to revise the functionality and description.

\subsubsection{Non-Destructive Output}
To maintain full data provenance, the system is non-destructive. The VLM's JSON response, which includes the \texttt{revised functionality}, \texttt{revised description}, and a \texttt{revision rationale}, is not used to overwrite any existing files. Instead, it is saved as a new, separate artifact, (e.g., \url{\{node\_id\}\_meta\_reannotated\_gemini-2.5-pro-thinking.json}).

This final file represents the culmination of our pipeline: a functional region defined by \textbf{human-verified geometry} and \textbf{VLM-revised semantics}, creating a high-fidelity data point for our benchmark.

\subsection{Functional Region Classification Details}
\label{sec:supp:region type cls}

The previous stages establish the geometric boundaries and functional descriptions of each region. However, to enable granular analysis and stratified evaluation (as seen in the experiments section, i.e., Sec.~\ref{sec:experiments}), each region must also be categorized into a standardized taxonomy (e.g., "Navigation Bar," "Modal," "Data Table"). This stage automates this classification process using \textbf{Gemini-2.5-Pro-Thinking}, ensuring consistency across the diverse GUI layouts in our dataset.

\subsubsection{Taxonomy Definition}
We define a comprehensive, hierarchical taxonomy tailored for modern GUIs, covering 10 high-level categories and over 40 specific types, shown in Tab.~\ref{tab:region type proportions}. This taxonomy is embedded directly into the Python script (\texttt{TAXONOMY} constant) and includes definitions such as:
\begin{itemize}
    \item \textbf{Primary Interface Containers:} Application Window, Browser Tab, Split-Screen Pane.
    \item \textbf{Global Navigation:} Header/Top Bar, Sidebar, Tab Bar, Breadcrumbs.
    \item \textbf{Content Display:} Main Content Area, Card List, Data Table, Dashboard Widget.
    \item \textbf{Interaction \& Input:} Search Region, Form, Filter Controls, Pagination.
    \item \textbf{Contextual Elements:} Modal/Dialog, Popover, Dropdown Menu, Notification Banner.
\end{itemize}
For leaf nodes (individual elements), the taxonomy is extended with types like "Button," "Link," and "Fragmented Element" to handle fine-grained classification.

\subsubsection{Classification Pipeline}
The classification is performed by a dedicated script (\texttt{classify\_functional\_regions.py}) that iterates through all annotated regions. The process for each region is as follows:

\begin{enumerate}
    \item \textbf{Filtering:} The script first filters out the root node ("Entire GUI") and any nodes that do not meet minimal size criteria, as these are trivial or non-informative.
    \item \textbf{Context-Aware Prompting:} The VLM is presented with the \textit{cropped image} of the region. To ensure accuracy, the prompt (\texttt{build\_classification\_prompt}) provides the model with:
    \begin{itemize}
        \item The full taxonomy list with definitions.
        \item A set of few-shot examples demonstrating correct classification (e.g., distinguishing an "Application Window" containing a dialog from the "Modal" itself).
        \item Instructions to select the \textit{single best fitting type} or use "Other" with a generated subtype if necessary.
    \end{itemize}
    \item \textbf{Robust Parsing:} The VLM's response is parsed to extract the \texttt{type}, \texttt{subtype}, and a \texttt{confidence} score. The system includes robust regex-based fallback mechanisms to handle varied VLM output formats.
    \item \textbf{Inheritance Logic:} For regions that are not explicitly classified (e.g., intermediate grouping nodes), the system implements a fallback inheritance logic (\texttt{get\_inherited\_type}), allowing them to adopt the classification of their parent or child nodes where appropriate to maintain semantic continuity in the hierarchy.
\end{enumerate}

\subsubsection{Output Integration}
The classification results are saved as separate JSON artifacts (e.g., \texttt{[node\_id]\_region-type.json}) alongside the node metadata. This modular approach allows the classification layer to be updated or refined independently of the core annotation data. These types are subsequently used to generate the breakdown statistics in our experimental results, allowing us to analyze model performance across different GUI components (e.g., distinguishing performance on "Navigation" vs. "Content" regions).

\subsection{More Dataset Statistics}
\label{sec:supp:data stats}

\subsubsection{Cost Analysis}
\textbf{Functional Region Division Cost.} The functional region division stage recursively detects functional regions, so accurate cost calculation is nearly infeasible. To provide an estimate, we simply consider the input size to always be $1920\times1080$. Using Gemini-2.5-Pro to detect the functional regions for the root image ($1920\times1080$), the average number of input tokens $N_{input}=4500$ plus that of output tokens $N_{output}=4000$ leads to $4500/1000000*1.25+4000/1000000*10=0.045625$ dollars\footnote{The Gemini-2.5-Pro pricing is listed at \url{https://ai.google.dev/gemini-api/docs/pricing\#gemini-2.5-pro}}. If the maximum of $10$ regions are detected at the 2nd level, the estimated cost detection is $10*(3000/1000000*1.25+4000/1000000*10)=0.4375$.

\textbf{Regional Refinement Cost.} Assuming one detected functional region is $1/10$ of the size of the root image ($1920\times1080$), the cost of one round of refinement with an average input tokens $1450$ and output tokens $480$ is $480/1000000*1.25+1450/1000000*10\approx0.0151$ dollars. If the maximum repeat of refinement $3$ is reached, then 10 regions cost $3*10*0.0151 \approx 0.453$ dollars.

\textbf{Re-Annotation Cost.} Still, assuming one detected functional region is $1/10$ of the size of the root image ($1920\times1080$), the cost of re-annotating a region with an average input tokens $1080$ and output tokens $1680$ is $1080/1000000*1.25+1680/1000000*10\approx0.01815$ dollars. Then 10 regions cost $10*0.01815 \approx 0.1815$ dollars.

The total cost is estimated to be $0.4375+0.453+0.1815=1.072$ dollars for processing one screenshot of size $1920\times1080$.

\subsubsection{Task Statistics}

\begin{table*}[]
\centering
\caption{The proportions of the types of the detected functional regions.}
\label{tab:region type proportions}
\resizebox{\textwidth}{!}{%
\begin{tabular}{@{}ccc@{}}
\toprule
\textbf{Top-Level Region Types}                    & \textbf{Secondary Region Types}                                                                                                                                                                                    & \textbf{Proportion (\%)} \\ \midrule
\rowcolor[HTML]{EFEFEF} 
Primary Interface Containers                      & Application Window, Browser Window / Tab, Split-Screen Pane                                                                                                                                                       & 16.4                \\
Global Navigation \& Structure                    & \begin{tabular}[c]{@{}c@{}}Header / Top Bar, Footer, Sidebar / Side Navigation, Tab Bar,\\ Toolbar / Action Bar, Breadcrumbs, Status Bar\end{tabular}                                                             & 54.3                \\
\rowcolor[HTML]{EFEFEF} 
Content \& Data Display                           & \begin{tabular}[c]{@{}c@{}}Main Content Area, Card / Item List, Dashboard / Widget Area,\\ Data Table / Grid, Image Gallery / Carousel, Media Player\end{tabular}                                                 & 11.1                \\
Interaction \& Input                              & \begin{tabular}[c]{@{}c@{}}Search Region, Form, Filter / Sort Controls, Comment Section,\\ Pagination Controls, Input field\end{tabular}                                                                          & 5.4                 \\
\rowcolor[HTML]{EFEFEF} 
Contextual \& Temporary Regions                   & \begin{tabular}[c]{@{}c@{}}Modal / Dialog Box, Popover / Tooltip, Dropdown Menu,\\ Context Menu, Notification / Toast / Alert Banner, Cookie Consent Banner\end{tabular}                                          & 9.4                 \\
Purely Static Content                             & Body Text, Static Title or Heading                                                                                                                                                                                & 0.8                 \\
\rowcolor[HTML]{EFEFEF} 
Individual Element                                & Button, Link, Image                                                                                                                                                                                               & 0.2                 \\
\multicolumn{1}{c}{System and Browser Artifacts} & \multicolumn{1}{c}{Scrollbar}                                                                                                                                                                                    & 0.2                 \\
\rowcolor[HTML]{EFEFEF} 
Other / Unknown                                   & \begin{tabular}[c]{@{}c@{}}Taskbar Item, Isolated Icon, Attachment Bar, User Profile Header,\\ Mobile Home Screen, Text Fragment, Logo,\\ Color Picker, Number Input with Stepper, Table of Contents\end{tabular} & 2.2                 \\ \bottomrule
\end{tabular}%
}
\end{table*}

\begin{figure}[t]
  \centering
   \includegraphics[width=1.0\linewidth]{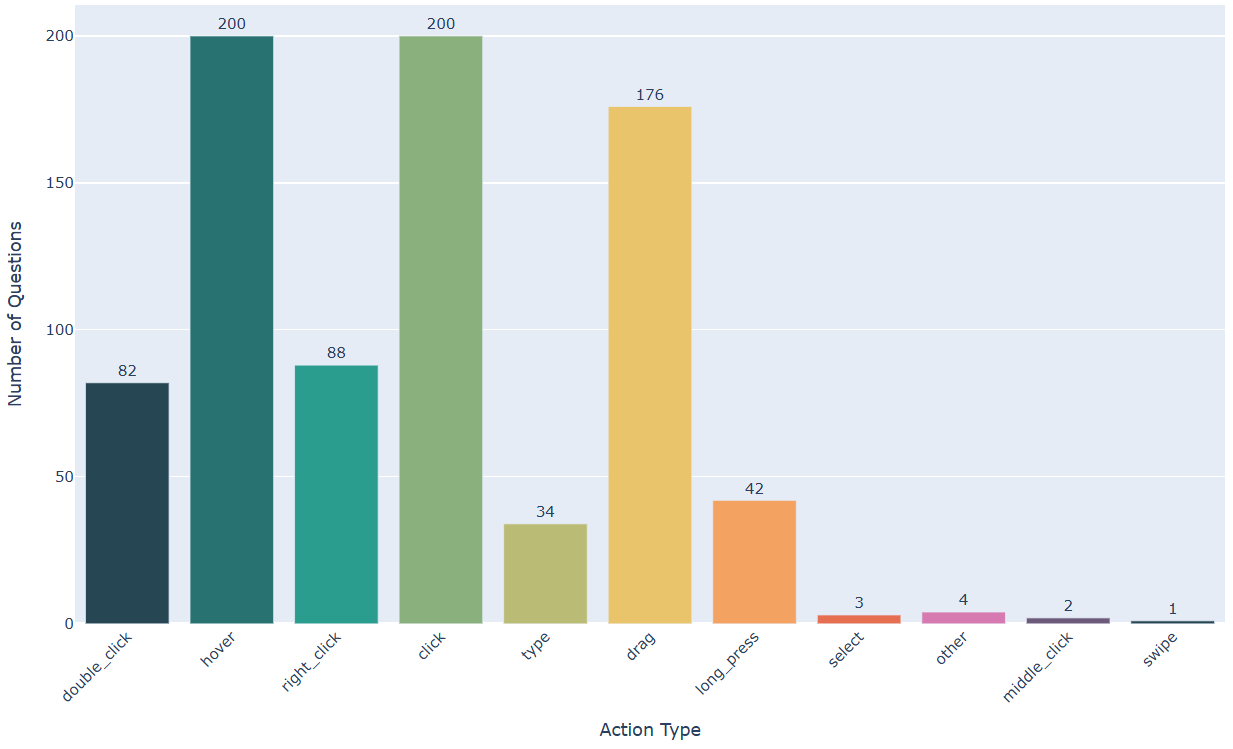}

   \caption{The number of the action types involved in the functionality-based element grounding tasks generated in Sec.~\ref{sec:element task gen}.}
   \label{fig:elemgnd acttype prop}
\end{figure}

\begin{figure}[t]
  \centering
   \includegraphics[width=1.0\linewidth]{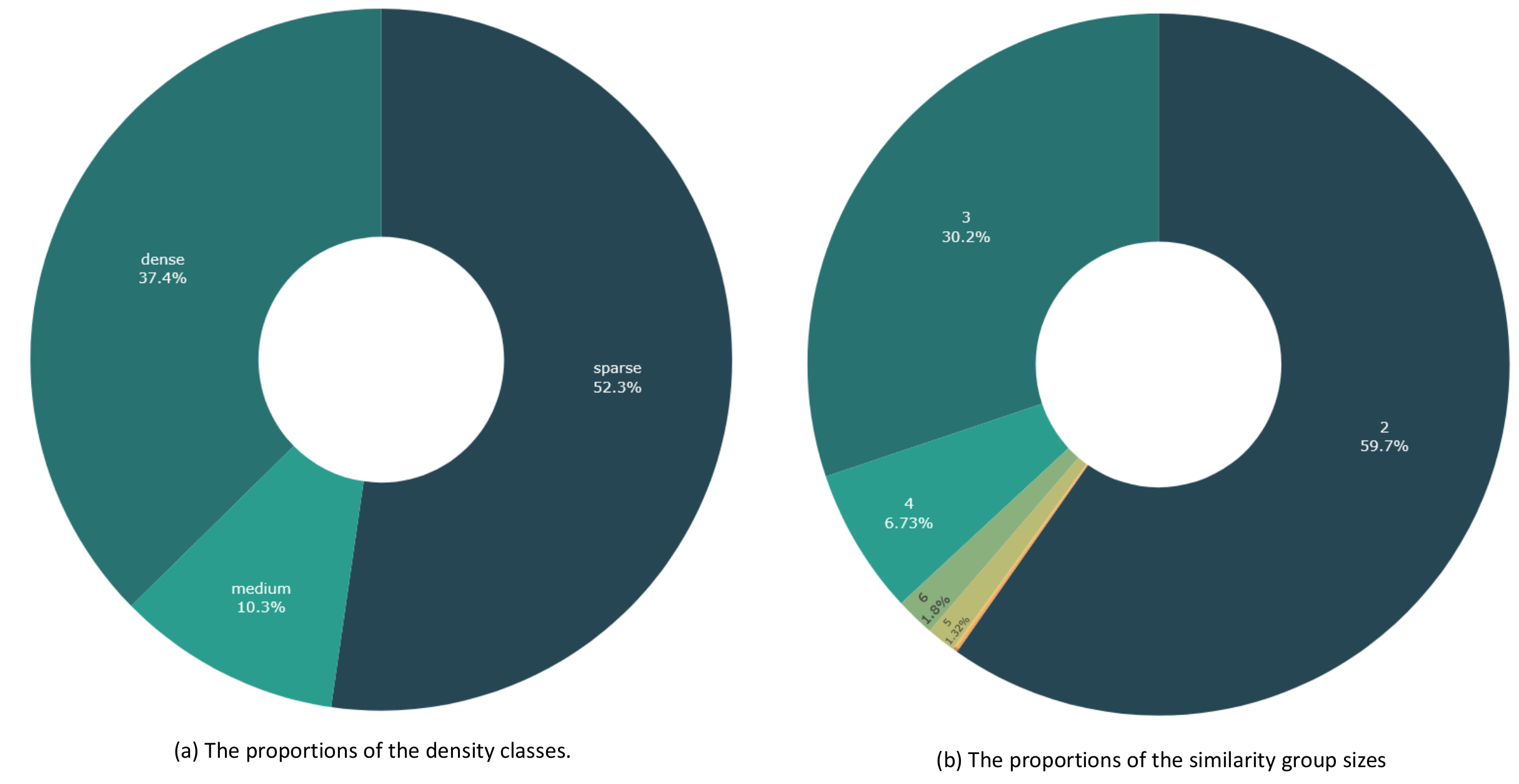}

   \caption{The proportions of the density classes and the similarity group sizes of the functionality-based element grounding tasks generated in Sec.~\ref{sec:element task gen}.}
   \label{fig:elemgnd pie charts}
\end{figure}

\begin{figure}[t]
  \centering
   \includegraphics[width=1.0\linewidth]{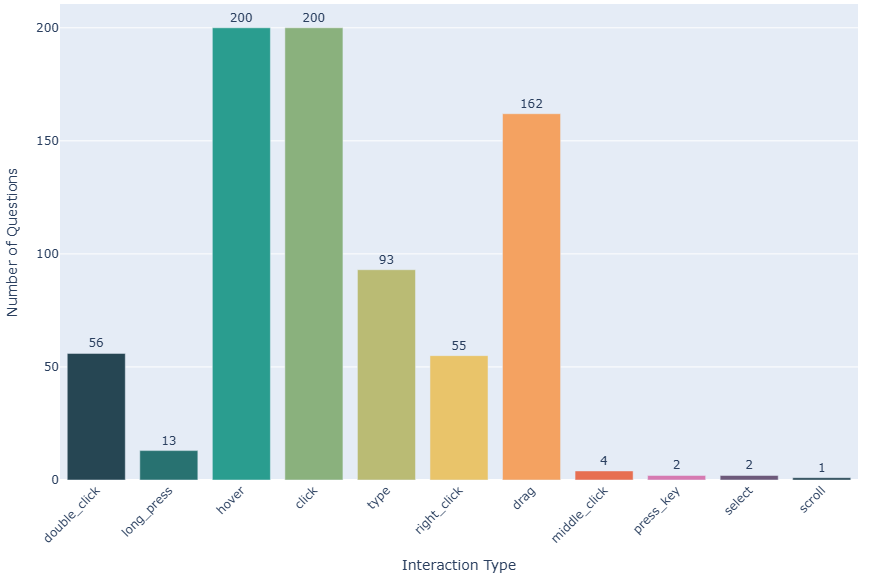}

   \caption{The number of the action types involved in the functionality-based element captioning tasks generated in Sec.~\ref{sec:element task gen}.}
   \label{fig:elemcap acttype prop}
\end{figure}

\begin{figure}[t]
  \centering
   \includegraphics[width=1.0\linewidth]{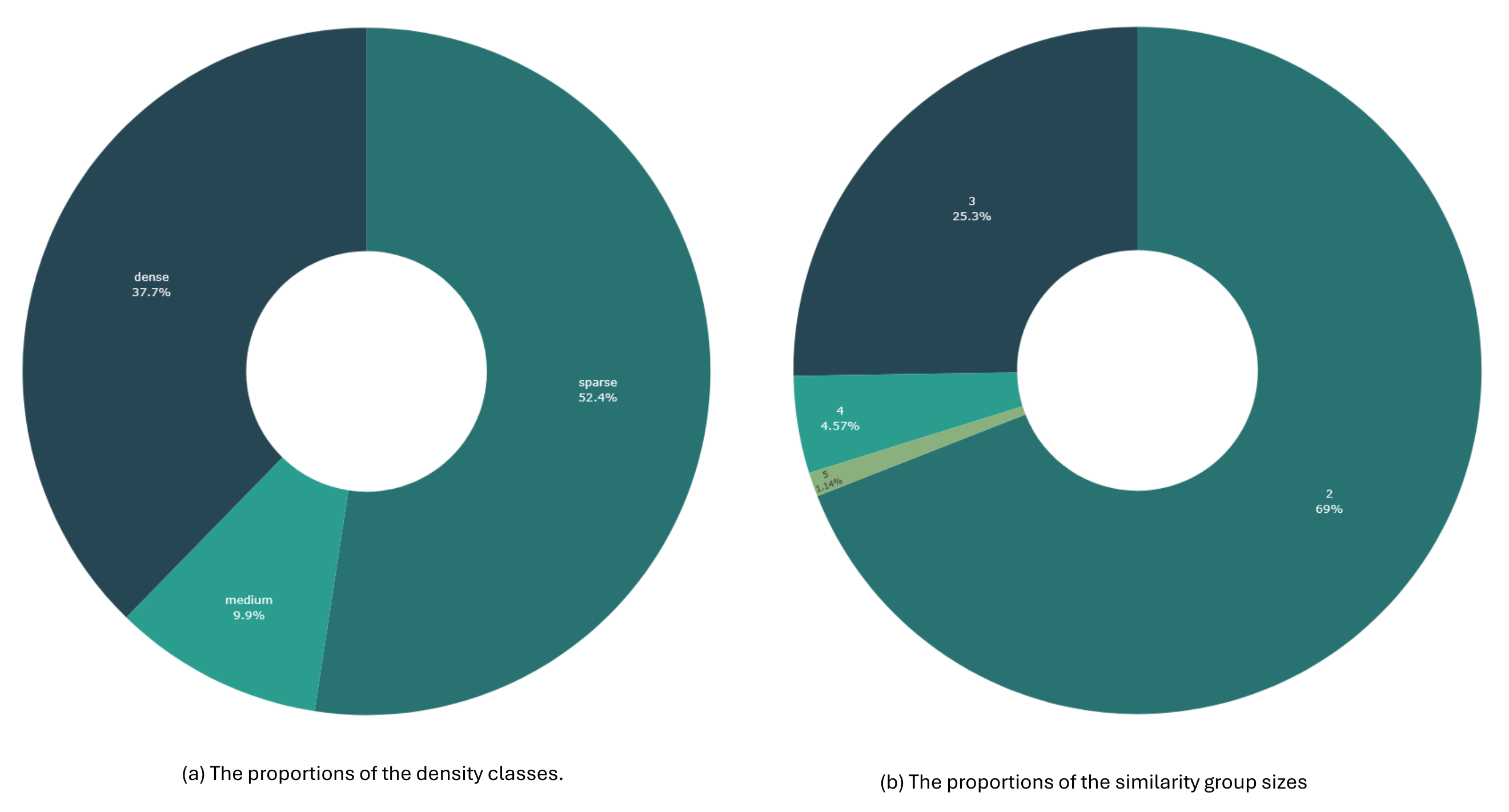}

   \caption{The proportions of the density classes and the similarity group sizes of the functionality-based element captioning tasks generated in Sec.~\ref{sec:element task gen}.}
   \label{fig:elemcap pie charts}
\end{figure}

The proportions of the region types are listed in Tab.~\ref{tab:region type proportions}. The task attributes (i.e., the action type involved, density class, and the size of the similarity group the target belongs to) are illustrated in Fig.~\ref{fig:elemgnd acttype prop}, Fig.~\ref{fig:elemgnd pie charts}, Fig.~\ref{fig:elemcap acttype prop}, and  Fig.~\ref{fig:elemcap pie charts}.


\subsection{Similar Functional Region Grouping Details}
To ensure that the generated evaluation tasks rigorously test an agent's ability to distinguish functionality rather than relying on superficial visual matching, we construct groups of visually similar but functionally distinct regions. This grouping process involves a cascade of semantic embedding, VLM-based visual verification, and deterministic geometric refinement.

\subsubsection{Initial Semantic Grouping.}
We utilize Qwen3-Embedding to compute embedding vectors for the visual descriptions of all functional regions. An affinity matrix is constructed based on cosine similarity. Regions are clustered into initial groups if their similarity exceeds a threshold of 0.6. To ensure the independence of elements within a group, we strictly filter out any pairs that exhibit a parent-child relationship using the hierarchy defined in \texttt{tree.json}. Furthermore, only regions that have undergone manual bounding box correction are considered for grouping to guarantee geometric precision.


\subsubsection{Visual Verification by VLM.}
Semantic embeddings of text descriptions may miss subtle visual nuances. Therefore, we employ Gemini-2.5-Pro-Thinking to visually inspect the initial groups against the original screenshot. The model performs three critical checks: (1) confirming that grouped elements are visually similar (e.g., in shape, color, or icon style) but perform different functions; (2) identifying potentially missing candidates from a provided list of other elements and adding them to the group, strictly ensuring that any added element does not overlap spatially (bounding box intersection) with existing group members; and (3) ensuring the group size remains within a reasonable range (2 to 5 elements).

\subsubsection{Geometric and Topological Processing.}
Following the visual verification, we apply a series of deterministic checks and refinements in Python:
\begin{enumerate}
    \item \textbf{Overlap Resolution:} We iteratively check for bounding box overlaps within each group. If an overlap is detected, one element (typically the one with the larger area) is removed, and the check is repeated until no overlaps remain.
    \item \textbf{Minimum Size Check:} Groups that have been reduced to fewer than 2 elements after verification or cleanup are immediately discarded.
    \item \textbf{Oversized Group Detection:} Groups containing more than 5 elements are flagged as ``oversized.''
    \item \textbf{Duplicate Merging:} We identify groups that share common elements. If two groups share $\ge 2$ elements, they are merged into a single cluster. During this merger, we re-apply the parent-child filter to remove any hierarchical conflicts introduced by the combination.
\end{enumerate}

\subsubsection{Refinement of Abnormal Groups.}
Groups flagged as ``oversized'' or those resulting from the merging process undergo a second pass processing by Gemini-2.5-Pro-Thinking. In this step, the model is tasked with selecting the optimal subset of elements (specifically targeting a size of 2 to 5) that best represents the visual pattern while maximizing functional confusion. Crucially, this step is subtractive only; the model can remove elements to refine the group but is not permitted to add new ones.

\subsubsection{Final Validation.}
A final Python-based validation is performed to ensure strict adherence to benchmark standards. We re-verify that no bounding box overlaps exist and that the final group size is strictly between 2 and 5 elements. Only groups passing these final constraints are used for task generation.

\section{Extra Evaluation Details}
\label{sec:supp:eval details}

\subsection{Functionality-Based Region Grounding Failure Case Inspection}

\begin{figure*}[th]
  \centering
   \includegraphics[width=0.85\linewidth]{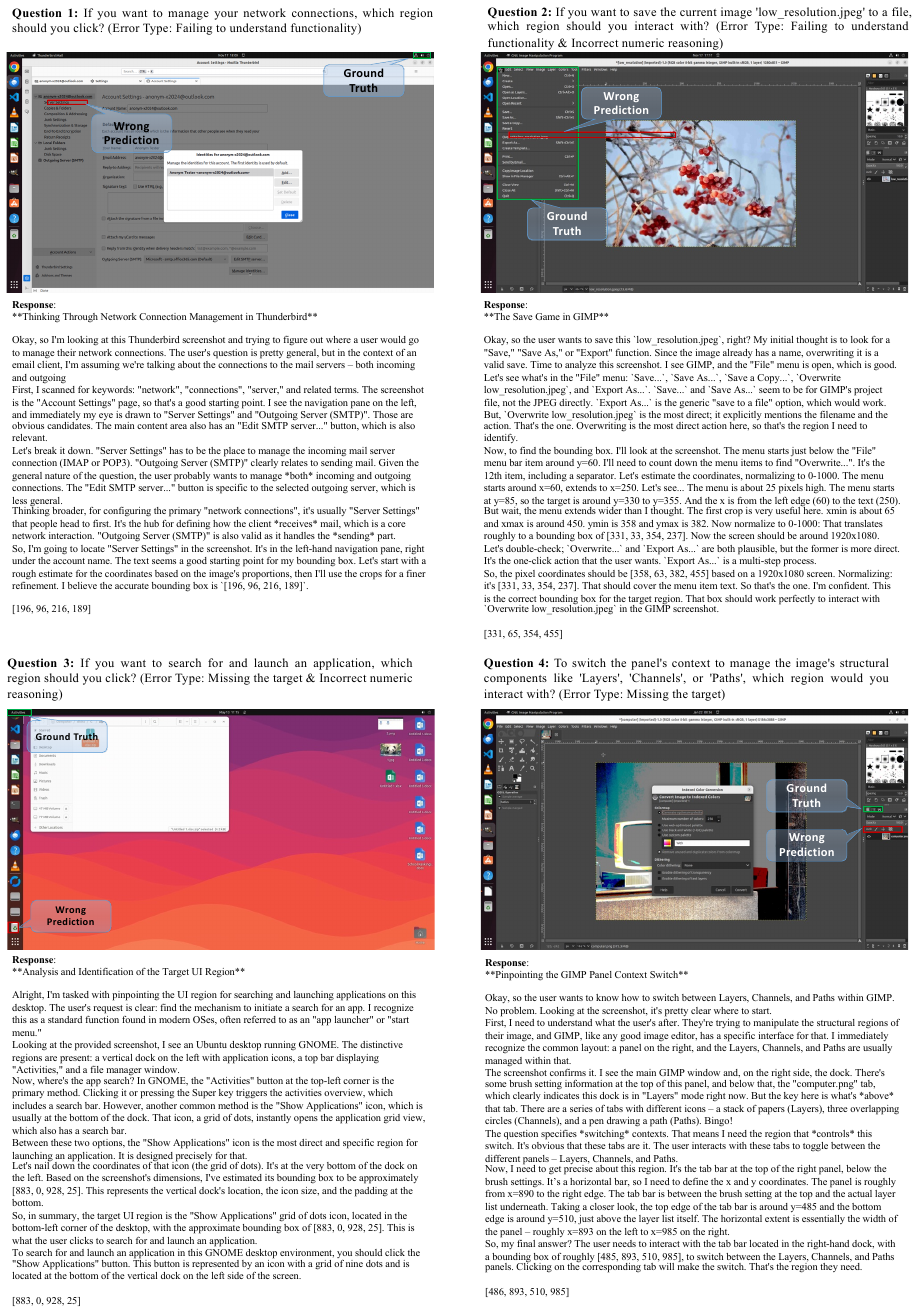}

   \caption{Visualization of typical failure cases by Gemini-2.5-Pro-Thinking on the functionality-based region grounding task.}
   \label{fig:regiongnd failure cases}
\end{figure*}
\begin{figure*}[th]
  \centering
   \includegraphics[width=0.85\linewidth]{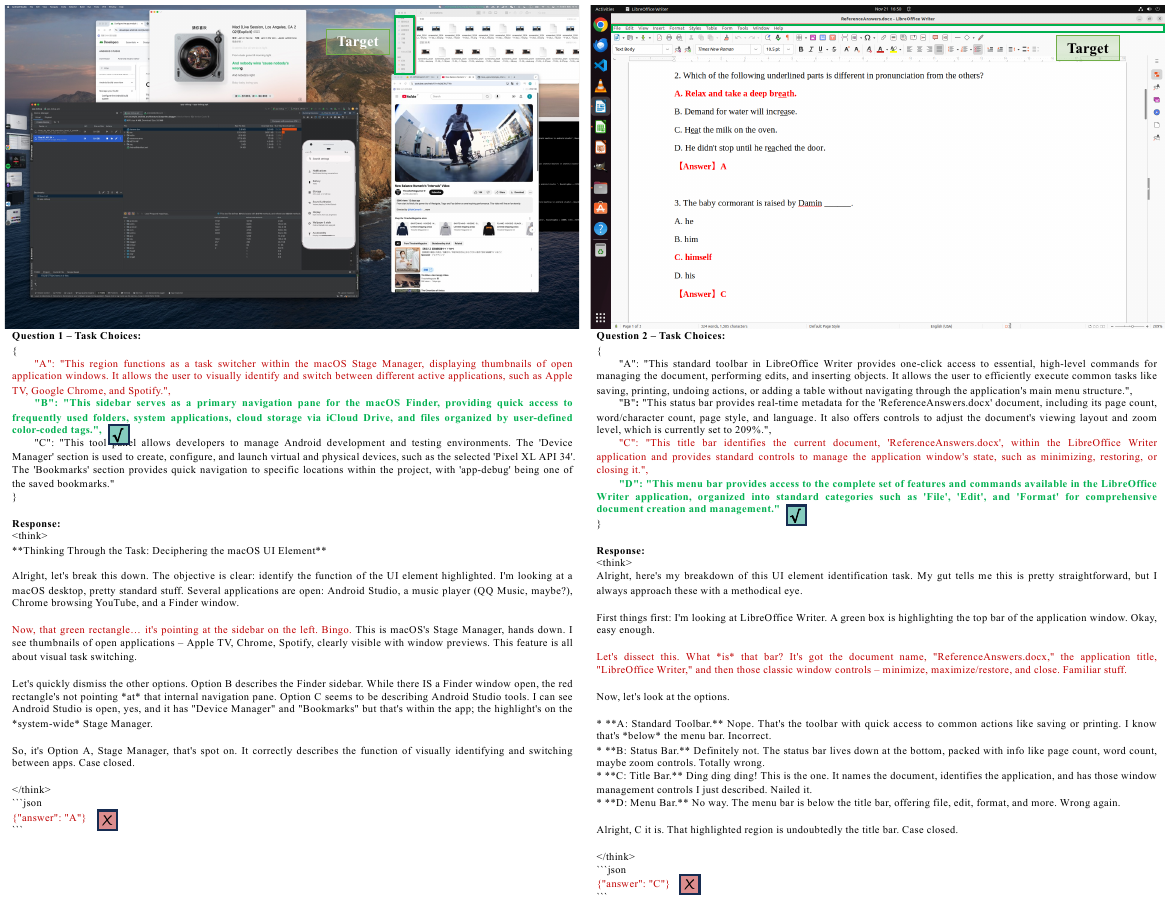}

   \caption{Visualization of typical failure cases by Gemini-2.5-Pro-Thinking on the functionality-based region captioning task.}
   \label{fig:regioncap failure cases}
\end{figure*}

In the grounding task, distinct failure patterns emerged that highlight a trade-off between semantic reasoning and localization precision. Gemini-2.5-Pro-Thinking demonstrates strong reasoning capabilities, often correctly identifying the target UI element corresponding to the functional description and locating the rough area. However, as shown in Fig.~\ref{fig:regiongnd failure cases}, its failure cases are predominantly characterized by spatial imprecision, where the predicted bounding boxes exhibit a low IoU with the Ground Truth. The generated boxes are frequently too loose or significantly shifted, despite covering the correct semantic region. Conversely, Qwen3-VL-32B-Instruct displays superior capabilities in bounding box regression, producing tight and geometrically accurate boxes. However, it falters significantly in target identification, frequently predicting high-confidence bounding boxes for the wrong elements. This indicates that while Gemini struggles with fine-grained pixel-level regression, Qwen tends to prioritize visual saliency over strict alignment with the functional textual description.

\subsection{Functionality-Based Region Captioning Failure Case Inspection}
For the captioning task, the error distribution reveals a shared deficiency in understanding the hierarchical structure of user interfaces. The primary failure mode for both Gemini-2.5-Pro-Thinking and Qwen3-VL-32B-Instruct falls into two main categories: differentiating between "System vs. Application" layers (e.g., confusing status bars with app-specific headers) and distinguishing different functionalities within a "Single Application" (e.g., misinterpreting similar icons within the same app context). Fig.~\ref{fig:regioncap failure cases} provides a detailed visualization of these failure patterns observed in Gemini. In addition to these shared errors, Qwen3-VL-32B-Instruct exhibits sporadic instances of "System vs. System" and "Inter-App" confusions. Collectively, these failure cases validate our hypothesis that existing models still lack a robust capability to predict and distinguish complex GUI functionalities, relying heavily on surface-level visual features rather than deep interactive logic.

\subsection{Functionality-Based Element Grounding Failure Case Inspection}

\begin{figure*}[th]
  \centering
   \includegraphics[width=0.85\linewidth]{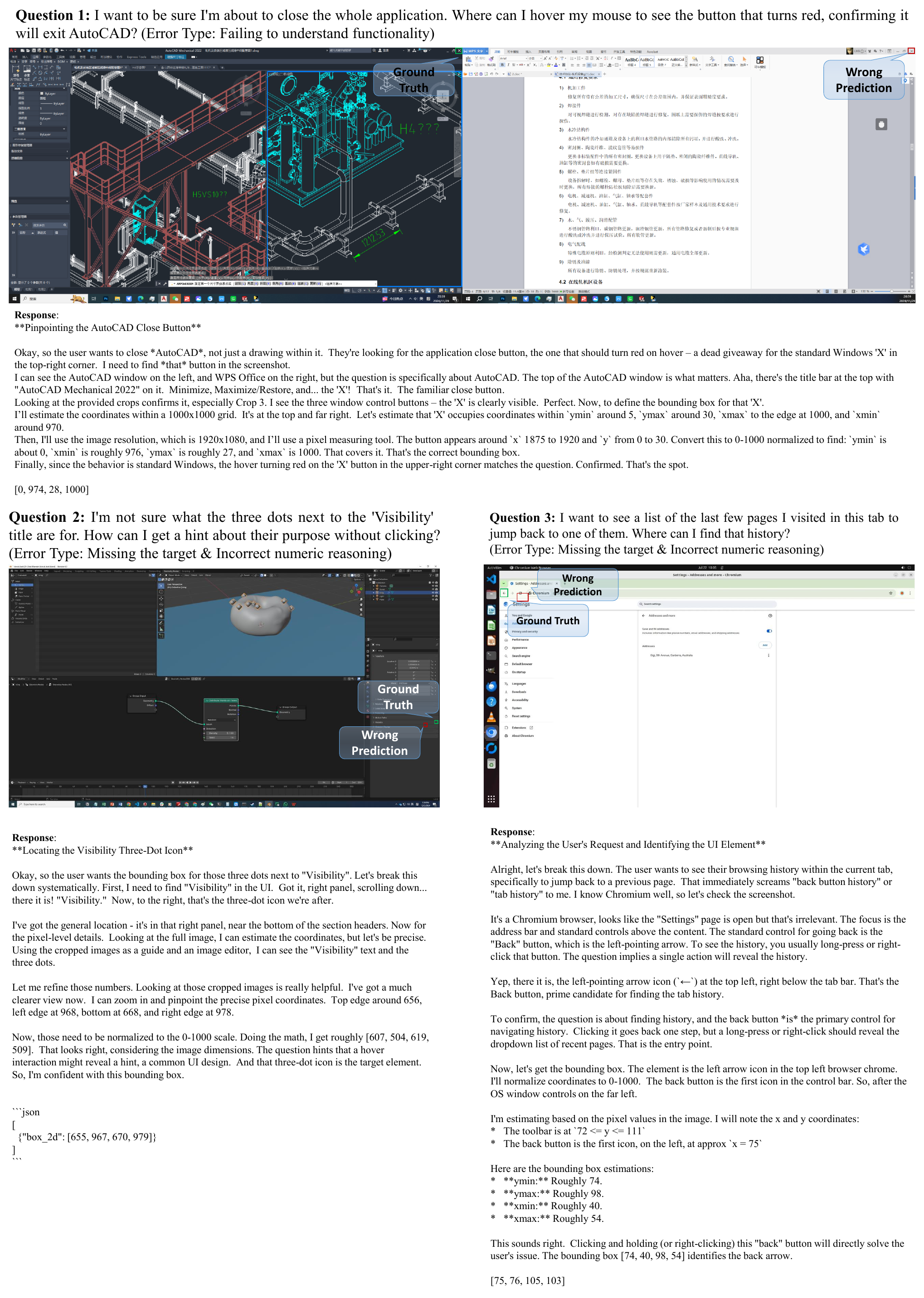}

   \caption{Visualization of typical failure cases by Gemini-2.5-Pro-Thinking on the functionality-based element grounding task.}
   \label{fig:elemgnd failure cases}
\end{figure*}
\begin{figure*}[th]
  \centering
   \includegraphics[width=0.85\linewidth]{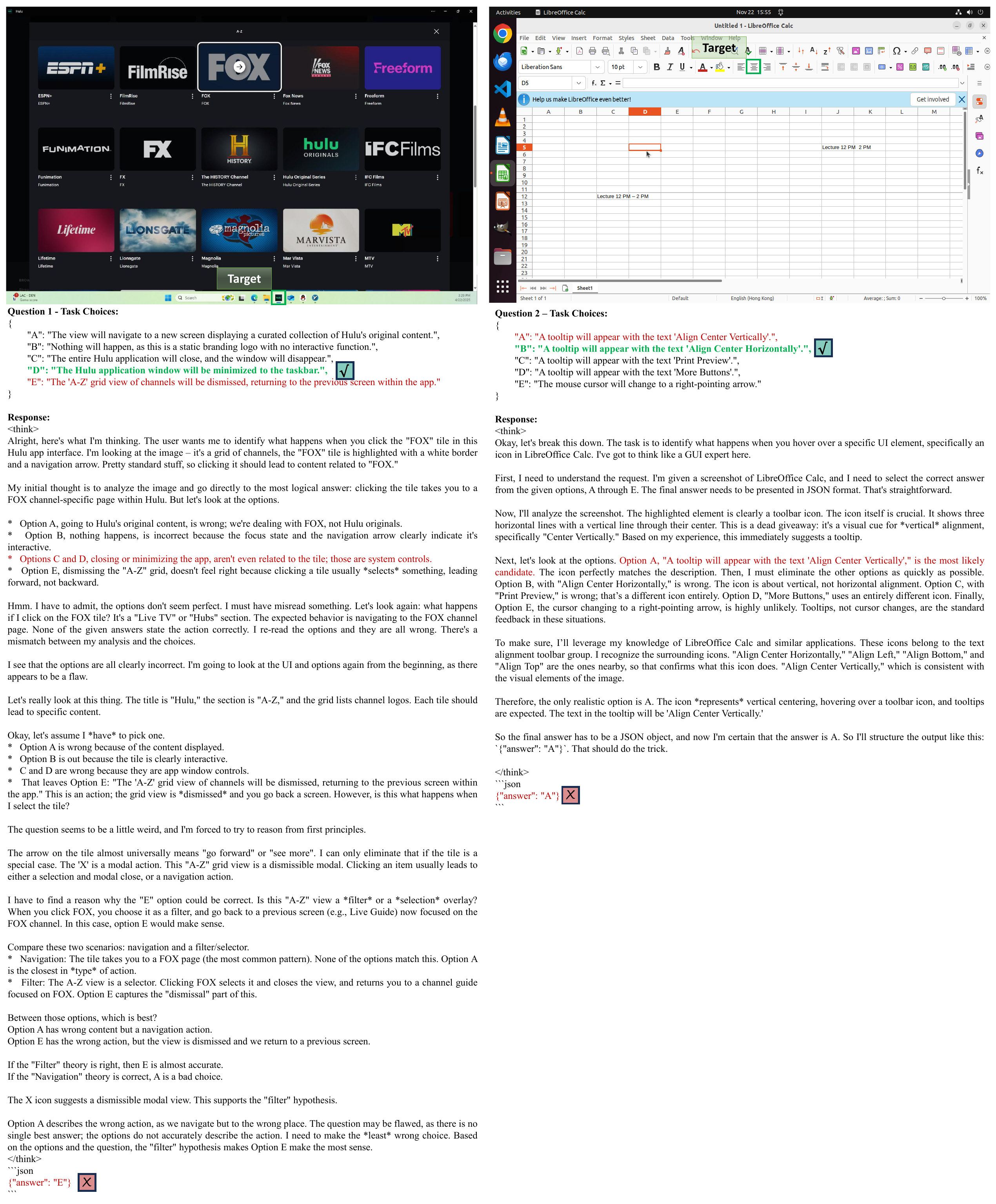}

   \caption{Visualization of typical failure cases by Gemini-2.5-Pro-Thinking on the functionality-based element captioning task.}
   \label{fig:elemcap failure cases}
\end{figure*}

To further investigate how the VLMs fail in the functionality-based element grounding, we manually inspected 30 cases and found that Gemini-2.5-Pro-Thinking owns the capability of pinpointing the target element in its reasoning content, but it cannot correctly predict the bounding box coordinates. For example, the reasoning content demonstrates redundant numeric computation about the four components of a box coordinate but is totally incorrect and useless. 90\% of cases are due to incorrect reasoning, while 10\% are due to a misunderstanding of functionality. Typical incorrect cases are visualized in Fig.~\ref{fig:elemgnd failure cases}.

\subsection{Functionality-Based Element Captioning Failure Case Inspection}

To further investigate how the VLMs fail in the functionality-based element captioning, we manually also inspected 30 cases and found that Gemini-2.5-Pro-Thinking also performs poorly. For example, Fig.~\ref{fig:elemcap failure cases} shows that this model fails to predict the outcome of clicking the app icon on the system tray at the bottom, mistakenly believing that this interaction just returns to the previous screen with the app. However, the real outcome is that the entire app window will be minimized. Moreover, this model also fails to distinguish the different alignment buttons in the Linux spreadsheet software. These examples indicate that even a powerful VLM like Gemini-2.5-Pro lacks satisfactory GUI functionality understanding capability.

\section{Prompt Suites}
\subsection{AutoGUI-v2 Annotating Prompts}
The prompts used by AutoGUI-v2 annotation pipeline are listed in Tab.~\ref{tab:division prompt}, Tab.~\ref{tab:checking prompt}, and Tab.~\ref{tab:reanno prompt}.

\onecolumn
\begin{longtable}{p{\dimexpr \textwidth - 2\tabcolsep}}

    \caption{The prompt used to instruct Gemini-2.5-Pro-Thinking to perform hierarchical GUI functional region division.} 
    \label{tab:division prompt} \\
    \toprule
    \textbf{Input Prompt Template} \\
    \midrule
    \endfirsthead
    
    \multicolumn{1}{c}{{\bfseries \tablename\ \thetable{} -- continued from previous page}} \\
    \toprule
    \endhead

    \midrule
    \multicolumn{1}{r}{{Continued on next page...}} \\
    \endfoot

    \bottomrule
    \endlastfoot

    
<image> You are an expert UI/UX analyst. Your task is to analyze a graphical user interface (GUI) screenshot and identify all functional regions. A functional region is a high-level grouping of individual UI elements (such as buttons, input fields, links, navigation elements, or icons) that collectively serve a single, overarching purpose.

Instructions:

1. Identify All First-Level Functional Regions: Scrutinize the entire GUI to identify every distinct, interactive region, including navigation bars, user profile management region, search regions, detailed content regions, sidebars, app windows, and other interactive regions.

2. Determine Functionality: For each identified region, first describe its visual layout and then deduce its primary functionality in detail.

- 2.1. Provide a high-level description of the region/element's function. Avoid detailing every specific functionality. Instead, focus on its broader impact on the webpage experience. For example, if interacting with a "Products" region reveals a dropdown menu, do not catalog the subsequent webpage changes in exhaustive detail.
    
- 2.2. To ensure uniqueness, your functionality description should reflect the instance-specific context of the region whenever possible. For example, instead of predicting `This region is used to search,' you should predict `This region allows users to search for electronic products on Amazon,' where `electronic products on Amazon' is specific to the current instance. Similarly, rather than predicting `This element facilitates the selection of an hour for the return time,' you should predict `This element updates the return time to 13 p.m on the clock picker.' if such information is directly available. Ensure that the description remains accurate, grounded in visible data, and does not speculate on unseen values.
    
3. Generate Normalized Bounding Boxes: For each region, provide the precise bounding box coordinates in a normalized format $[x_{min}, y_{min}, x_{max}, y_{max}]$ . The coordinates must be between 0 and 1000, where $(x_{min}, y_{min})$ is the top-left corner and $(x_{max}, y_{max})$ is the bottom-right corner of the region. All boxes should tightly bound the region and MUST not overlap.

4. Be Thorough: Identify all non-overlapping regions. Skip purely decorative or non-interactive elements like static text, background images, or simple separators.

5. Division Limit: The number of the detected functional regions should be less than 10 as too many regions may indicate that you accidentally break down a large region into too many small regions.

6. Determine Dividability: For each region, determine whether it can be divided into smaller lower-level self-contained functional regions. If yes, set the "dividable" field to true; if no, set it to false.

7. Classify the types of the functional regions according to the following dictionary:

\{
    "Primary Interface Containers": \{\\
        "Entire GUI": "Only used for the overall GUI item",\\
        "Application Window": "Main container for a desktop application, including window controls and all internal UI.",\\
        "Browser Window / Tab": "Container for a single webpage within a browser, including the address bar and tab UI.",\\
        "Split-Screen Pane": "A divided section of a window for displaying multiple views or documents simultaneously."\\
    \},\\
    "Global Navigation \& Structure": \{\\
        "Header / Top Bar": "Top-most region with the logo, main navigation, search, and account access.",\\
        "Footer": "Bottom-most region with secondary links, copyright, and contact info.",\\
        "Sidebar / Side Navigation": "Vertical panel for navigation, content hierarchy (e.g., a file tree), or filters.",\\
        "Tab Bar": "A set of tabs to switch between different views, sections, or documents.",\\
        "Toolbar / Action Bar": "A set of controls or icon buttons for performing common actions.",\\
        "Breadcrumbs": "Navigation trail showing the user's current location in the UI hierarchy."\\
    \},\\
    "Content \& Data Display": \{\\
        "Main Content Area": "Primary region for displaying main content like an article, video, or document.",\\
        "Card / Item List": "A list or grid of repeating items (cards), such as products or social media posts.",\\
        "Dashboard / Widget Area": "A summary view of data, metrics, and visualizations presented as widgets.",\\
        "Data Table / Grid": "Displays data in a sortable and filterable table with rows and columns.",\\
        "Image Gallery / Carousel": "An interactive viewer for a collection of images or promotional banners.",\\
        "Map View": "An interactive map for displaying geographical data.",\\
        "Media Player": "Region for playing video or audio with playback controls."\\
    \},\\
    "Interaction \& Input": \{\\
        "Search Region": "An input field and button for performing a search.",\\
        "Form": "A set of fields for user data submission (e.g., registration, contact).",\\
        "Filter / Sort Controls": "Controls for filtering, refining, and sorting content.",\\
        "Login / Authentication Form": "A specific form for user login with username and password fields.",\\
        "Comment Section": "Region for users to read and write comments.",\\
        "Pagination Controls": "Controls (e.g., page numbers, 'next'/'previous') to navigate paged content."\\
    \},\\
    "Contextual \& Temporary Regions": \{\\
        "Modal / Dialog Box": "A pop-up overlay that requires user interaction to be dismissed.",\\
        "Popover / Tooltip": "A small overlay that shows extra information on hover or click.",\\
        "Dropdown Menu": "A list of options that appears when an element is clicked.",\\
        "Context Menu": "A menu of relevant actions that appears on right-click.",\\
        "Notification / Toast / Alert Banner": "A temporary message that provides feedback or status updates.",\\
        "Cookie Consent Banner": "A banner that informs users about cookies and asks for their consent."\\
    \},\\
    "Others (You should come up with standard type names you think are most suitable for the region)": \{\\
        "Others (Come up with a sub-type name, too)": "Other functional regions that you believe are a better fit but do not fall into the above categories."\\
    \}\\
\}

8. **Adhere to the Output Format Below:** Provide the final output as a single JSON array. Please first describe the overall GUI and then the detected functional regions in fine detail. Do not hallucinate non-existing visual contents. Set the "type" field of the overall GUI item to "Entire GUI".
Output Format:

\verb|```|json

[

  \{\\
    "id": ".../10" (The maximum number of regions is 10),\\
    "bbox": ...,\\
    "dividable": (true means this region can be divided into smaller self-contained functional regions, false means this region cannot be divided anymore),\\
    "type": (The type of the functional region),\\
    "description": "(Region layout and appearance description in English)",\\
    "functionality": "(Region's contextual functionality description in English)"\\
  \},\\
  ...

]

\verb|```|\\

If the screenshot represents an individual element, just mention the overall GUI as it cannot be divided into smaller regions.

6. Here are 4 examples

Example 1: Given a screenshot of VS Code running on Ubuntu, you can output the following JSON:

[\\
  \{\\
    "id": "1/10",\\
    "bbox": [0, 0, 1000, 1000],\\
    "dividable": true,\\
    "type": "Entire GUI",\\
    "description": "A screenshot of a Ubuntu desktop environment, with the Visual Studio Code application open as the main window. The application is in dark mode and is displaying the settings page.",\\
    "functionality": "The user interface provides access to the operating system's features and applications, with a primary focus on code editing and project management within the Visual Studio Code editor."\\
  \},\\
  \{\\
    "id": "2/10",\\
    "bbox": [231, 197, 766, 946],\\
    "dividable": true,\\
    "type": "Application Window",\\
    "description": "A prominent rectangular window located in the center of the desktop, with a distinct dark theme and a menu bar at the top.",\\
    "functionality": "This is the primary application window for Visual Studio Code, providing an integrated development environment for editing code, managing project files, and accessing development tools."\\
  \},\\
  \{\\
    "id": "3/10",\\
    "bbox": [0, 0, 1000, 22],\\
    "dividable": true (Because this top bar comprises these functional regions: Activities Button, Active Application Menu, Clock and Notification Center, and System Status Menu),\\
    "type": "Header / Top Bar",\\
    "description": "A horizontal bar that spans the entire width of the screen at the very top, featuring a black background and white text.",\\
    "functionality": "This top bar provides essential system-wide information and controls, including the current time and date, system status indicators, and access to system-level settings and notifications."\\
  \},\\
  \{\\
    "id": "4/10",\\
    "bbox": [0, 22, 31, 1000],\\
    "dividable": false (Because this dock only contains undividable elements, which do not form a functional region),\\
    "type": "Sidebar / Side Navigation",\\
    "description": "A vertical, semi-transparent bar located on the left side of the screen, containing a series of application icons.",\\
    "functionality": "This vertical dock serves as the application launcher for the Ubuntu operating system, providing a convenient way for users to quickly launch and switch between their favorite and running applications."\\
  \}\\
]

Example 2: Given a screenshot of MS Excel, with a modal pop-up window titled "Move/Copy Sheet" centered on the screen and being the only interactive region, you can output:

[\\
  \{
    "id": "1/10",\\
    "bbox": [0, 0, 1000, 1000],\\
    "dividable": true,\\
    "type": "Entire GUI",\\
    "description": "The user interface of the Excel, showing a spreadsheet titled \"NGSS Assessment Resource Hyperlink\". A \"Move/Copy Sheet\" modal dialog is open and centered, partially obscuring the spreadsheet data.",\\
    "functionality": "This interface allows for the creation, editing, and management of spreadsheet data. The currently active dialog provides functionality to reorganize worksheets within the workbook."\\
  \},\\
  \{\\
    "id": "2/10",\\
    "bbox": [390, 240, 608, 706],\\
    "dividable": true (Because this dialog box comprises Move/Copy Sheet Button, Destination Folder Button, and New Name Input Field),\\
    "type": "Modal / Dialog Box",\\
    "description": "A small, rectangular pop-up window centered on the main application interface, with a title bar labeled 'Move/Copy Sheet' at the top.",\\
    "functionality": "This dialog box allows users to move or copy a selected sheet within the current Excel spreadsheet, providing options to choose the action, destination, and new name for the sheet."\\
  \}\\
]\\

Example 3: Given a screenshot of a single search bar element on YouTube, you can output:

[\\
  \{\\
    "id": "1/10",\\
    "bbox": [0, 0, 1000, 1000],\\
    "dividable": false,\\
    "type": "Entire GUI",\\
    "description": "A horizontally elongated, pill-shaped input field with a light gray background. It contains the placeholder text 'Search' and a magnifying glass icon on the right side, indicating the search function.",\\
    "functionality": "This search bar is the primary tool for users to find and discover video content on the YouTube platform. Users can type keywords, phrases, or video titles to search for relevant videos, channels, or playlists."\\
  \}\\
]

Example 4: Given an incomplete or corrupted screenshot of a functional region that cannot be processed, you can output an empty JSON array:

[]

I have given you a \{is\_dividable\}GUI screenshot\{context\}.

Now it's your turn.

 \\
    \midrule

    \textbf{Response Example}\\
    \midrule

<think>
**Initial Assessment \& Region Identification**

Okay, first things first: I'm looking at a screenshot of an Ubuntu desktop. Clearly, the main element is a Visual Studio Code window open to the "Settings" tab, focusing on the "Files: Auto Save" option.  My initial thought is to break this down into logical regions.  I'm seeing: 1. the entire desktop itself (the overarching container), 2. the Ubuntu top bar, 3. the Ubuntu dock on the left, and 4. the VS Code window. Each of these have distinct visual and functional boundaries.

**Deep Dive \& Feature Analysis**

Let's dissect each region. The whole screen (region 1) gives us the Ubuntu interface, a VS Code window, and a generally dark aesthetic. That's the canvas. Functionally, it's about providing the OS and running an IDE. The top bar (region 2) is Ubuntu's control center, showing the active app, time, and system info. Its function is global navigation. Then there's the dock (region 3), a vertical app launcher on the left. Finally, VS Code (region 4) – the star of the show here – is showing the settings pane. I can tell this is dividable because it's got a file explorer, the settings content, and a title bar.

**Bounding Boxes \& Refinement**

Now, I'm thinking about bounding boxes. I need to get those BBoxes right. The full screen is easy, [0, 0, 1000, 1000]. I'll eyeball the top bar as [0, 0, 1000, 23]. The dock is a vertical strip on the left, so about [0, 23, 32, 1000].  The VS Code window starts roughly at [218, 197, 781, 946] -- pretty much spot on. The BBoxes feel correct.

**Final JSON Generation**

Okay, ready to structure this into a JSON.  Each item gets an ID, description, functionality, type, and importantly, that bounding box I calculated. I will construct the final JSON output based on this. I've got a description of each region -- what it is, what it does. The functionality part is critical -- it explains what the user is doing within that area. Finally, the "dividable" field -- important to note whether something can be further broken down. Alright, let's craft that output now.

</think>

\verb|```|json

[

  \{
    "id": "1/10",\\
    "bbox": [0, 0, 1000, 1000],\\
    "dividable": true,\\
    "type": "Entire GUI",\\
    "description": "A screenshot of the Ubuntu desktop environment. The main window is Visual Studio Code, which is open to the Settings page. A vertical dock with application icons is visible on the left, and a system status bar is at the top. The desktop background is a purple and orange geometric pattern.",\\
    "functionality": "This GUI provides a complete desktop experience for the Ubuntu operating system, allowing users to manage files, launch applications, and interact with the system, with the current focus on configuring the Visual Studio Code editor."\\
  \},
  
\{\\
    "id": "2/10",\\
    "bbox": [0, 0, 1000, 23],\\
    "dividable": true,\\
    "type": "Header / Top Bar",\\
    "description": "A black horizontal bar at the top of the screen displaying 'Activities', the current application name 'Visual Studio Code', the date and time 'Jan 22 03:31', and system status icons on the right.",\\
    "functionality": "This is the global top bar for the Ubuntu operating system, providing access to system-level functions, displaying the active application's name, and showing status information like time, date, and system notifications."\\
  \},\\
  \{\\
    "id": "3/10",\\
    "bbox": [0, 23, 32, 1000],\\
    "dividable": false,\\
    "type": "Sidebar / Side Navigation",\\
    "description": "A vertical bar on the left side of the screen containing a column of application icons, including Google Chrome, Thunderbird, Visual Studio Code, and others.",\\
    "functionality": "This dock serves as the primary application launcher and task manager for the Ubuntu desktop, allowing users to quickly start their favorite programs and switch between open windows."\\
 \},\\
 
  \{\\
    "id": "4/10",\\
    "bbox": [218, 197, 781, 946],\\
    "dividable": true,\\
    "type": "Application Window",\\
    "description": "The main application window for Visual Studio Code, featuring a dark theme. It has a title bar, a file explorer sidebar on the left showing the 'HW-8-MAIN' project, and a main content area displaying the 'Settings' tab, which is focused on the 'Files: Auto Save' option.",\\
    "functionality": "This window provides the integrated development environment (IDE) for Visual Studio Code. The user is currently interacting with the settings interface to configure the editor's behavior, specifically adjusting the auto-save functionality for files."\\
  \}\\
  
]

\verb|```| \\


\end{longtable}
\twocolumn
\onecolumn
\begin{longtable}{p{\dimexpr \textwidth - 2\tabcolsep}}

    \caption{The prompt used to instruct Gemini-2.5-Pro-Thinking to perform hierarchical GUI functional region division.} \label{tab:checking prompt} \\
    \toprule
    \textbf{Input Prompt Template} \\
    \midrule
    \endfirsthead

    \multicolumn{1}{c}{{\bfseries \tablename~\thetable{} -- continued from previous page}} \\
    \toprule
    \textbf{Input Prompt Template} \\
    \midrule
    \endhead

    \midrule
    \multicolumn{1}{r}{{Continued on next page...}} \\
    \endfoot

    \bottomrule
    \endlastfoot


<image> You are a meticulous AI assistant specializing in visual analysis and quality control. Your task is to evaluate a cropped image depicting a functional region within a larger GUI screenshot and determine the completeness and boundedness of the shown region, providing a quantifiable score.

Useful Guidelines:\\
A functional region is a high-level grouping of individual UI elements (such as buttons or input fields) that collectively serve a single, overarching purpose. In images with multiple overlapping windows, the primary application window in focus should be considered the main functional region.

A region box is considered valid if it meets both of the following criteria:\\
- Completeness: This measures how much of the functional region is visible and unobstructed within the provided boundaries. A region is considered incomplete if its key components are obscured or occluded by other overlapping elements (like pop-up dialogs or other windows). This score is independent of the box's framing.

- Boundedness: This measures the quality of the framing. A region is poorly bounded if the box is either: (a) Too tight: The box crops, clips, or cuts off any part of the functional region.\\
Too loose: (b) The box includes excessive, non-functional padding or background area that is not part of the subject.

Your Task\\
Analyze the user-provided image and its highlighted region. Then, provide your assessment in the specified format.

Scoring Rubrics\\
A. Completeness Score (Evaluates Occlusion):

3 (Perfect): The functional region is fully visible within the box. No part of it is occluded by other elements.\\
2 (Almost Complete): Only minor, non-critical parts of the region are occluded.\\
1 (Mostly Incomplete): A large or critical part of the region is occluded.\\
0 (Severely Incomplete): The vast majority of the functional region is occluded or missing.

B. Boundedness (Evaluates Framing): (a) Yes: The box tightly frames the entire functional region without cropping it and without including excessive padding. (b) No: The box either crops the functional region or includes excessive padding.

Critical Rule: If a region is cropped by its boundary (Boundedness: No), its Completeness Score cannot be 3, because a cropped view is inherently an incomplete one.

Instructive Examples:\\
Example 1: Perfect Case\\
(Image: A single application window perfectly filling the screenshot from edge to edge.)

Reasoning: The region is perfectly framed without being cropped, and its content is fully visible.\\
Completeness Score: 3\\
Boundedness: Yes

Example 2: Poor Boundedness (Excessive Padding)\\
(Image: A small calculator application window shown in the middle of a large, empty desktop.)

Reasoning: The region is fully visible, but it is poorly bounded due to excessive empty padding around it.\\
Completeness Score: 3\\
Boundedness: No

Example 3: Poor Boundedness (Cropping)\\
(Image: A web browser window where the right side and bottom are clearly cut off by the edge of the screenshot.)

Reasoning: The region is poorly bounded because it is cropped by the image frame, making the view incomplete.\\
Completeness Score: 1\\
Boundedness: No

Example 4: Poor Completeness (Occlusion)\\
(Image: A perfectly framed file explorer window, but a "Save As" dialog box is covering the central list of files.)

Reasoning: The region is well-bounded, but its completeness is low because a critical area is occluded by a dialog box.\\
Completeness Score: 1\\
Boundedness: Yes

Example 5: Poor Completeness AND Poor Boundedness\\
(Image: An image of VSCode, which is both occluded by a browser and cropped at the bottom.)

Reasoning: The region is poorly bounded because it is cropped, and its completeness is low due to significant occlusion from the overlapping browser window.\\
Completeness Score: 1\\
Boundedness: No

Output Format:\\
Reasoning: [Provide a brief, one-sentence explanation for the score, referencing incompleteness, excessive padding, or perfect framing.]\\
Completeness Score: [0-3]\\
Boundedness: [Yes/No]

I have given you an image of a functional region.\\
Now it's your turn.

    \\
    \midrule

    \textbf{Response Example}\\
    \midrule

<think>\\
Okay, here's my analysis. The user wants me to analyze this cropped GUI region, specifically the top bar of a likely GNOME desktop. I need to assess how well the cropped image represents the complete top bar.

First, I identified the functional region: it's the top bar with "Activities", application name, time, and system status icons.

Next, I looked at *Completeness*. The original image has the full desktop, and the cropped region is just the top bar. Nothing occludes the bar itself. So, it's fully *visible* at least.

Then, I assessed *Boundedness*. This is key. The red box in the original shows that the cropped region *does* cut off the right edge of the top bar, impacting system tray icons and likely other status indicators. This is significant cropping. Therefore, Boundedness is "No".

Now, I apply the crucial rule: "If Boundedness is No, Completeness cannot be 3." Since the top bar *is* cropped, I can't give it a perfect completeness score.

I need to decide what score to give. The left side is mostly there, but the right, with the system tray, is substantially cut off. That's a huge piece of functionality, the system status and controls, missing.  It's crucial! This means a score of 1 makes the most sense because a large portion of the area is missing.

So, here's my final output.

Reasoning: The region is poorly bounded because it is cropped on the right side, cutting off the system tray icons.\\
Completeness Score: 1\\
Boundedness: No\\

</think>\\
Reasoning: The region is poorly bounded because it is cropped on the right side, cutting off the system tray icons.\\
Completeness Score: 1\\
Boundedness: No

\end{longtable}
\twocolumn
\onecolumn
\begin{longtable}{p{\dimexpr \textwidth - 2\tabcolsep}}

    \caption{The prompt used to instruct Gemini-2.5-Pro-Thinking to re-annotate the functionality and layout description for each functional region.} \label{tab:reanno prompt} \\
    \toprule
    \textbf{Input Prompt Template} \\
    \midrule
    \endfirsthead

    \multicolumn{1}{c}{{\bfseries \tablename~\thetable{} -- continued from previous page}} \\
    \toprule
    \endhead

    \midrule
    \multicolumn{1}{r}{{Continued on next page...}} \\
    \endfoot

    \bottomrule
    \endlastfoot


<image> You are an expert UI/UX analyst. Your task is to analyze a graphical user interface (GUI) screenshot and revise the functional description for a GUI region after a human corrected its bounding box.\\

Context:\\
  • Previously annotated functionality: \{previous\_functionality\}\\
  • Previously annotated description: \{previous\_description\}\\

Look carefully at the red rectangle on the full screenshot and the cropped image of that region. Revise and polish the functionality and description of the specified region.\\

Requirements:\\
1. The previously annotated functionality and description may be incorrect, either containing hallucinated details or missing important discernible details. Therefore, you MUST correct the functionality and description according to the bounding box marked on the screenshot and the problems encountered in the original annotation.\\
2. You should revise the functionality according to the format requirements shown below.\\
2.1. The revised functionality MUST provide a high-level description of the region/element's function. Avoid detailing every specific functionality. Instead, focus on its broader impact on the webpage experience. For example, if interacting with a "Products" region reveals a dropdown menu, do not catalog the subsequent webpage changes in exhaustive detail.\\
2.2. To ensure uniqueness, your functionality description should reflect the instance-specific context of the region whenever possible. For example, instead of predicting 'This region is used to search,' you should predict 'This region allows users to search for electronic products on Amazon,' where 'electronic products on Amazon' is specific to the current instance. Similarly, rather than predicting 'This element facilitates the selection of an hour for the return time,' you should predict 'This element updates the return time to 13 p.m on the clock picker.' if such information is directly available. Ensure that the description remains accurate, grounded in visible data, and does not speculate on unseen values.\\
3. You should also revise the description. A description should describe the region's layout and appearance in English and in detail.\\

Respond strictly in JSON with this schema:\\
\verb|```|json\\
\{
  "revision rationale": "Carefully describe the whole GUI screenshot, the region marked with a red rectangle, and their relationship. Then, describe the rationale for the revision.",\\
  "revised description": "...",\\
  "revised functionality": "..."\\
\}\\
\verb|```|

Do not include any extra keys or commentary outside of the JSON object.

    \\
    \midrule

    \textbf{Response Example}\\
    \midrule

\verb|```|json\\
\{\\
  "revision rationale": "The user interface is a desktop environment, likely Ubuntu with the GNOME desktop, showing the LibreOffice Calc application as the active window. The red rectangle highlights the vertical dock on the left side of the screen. This dock contains pinned and running applications. The previously annotated functionality and description are generally correct but lack completeness. The description fails to mention several key elements, most notably the 'Show Applications' grid icon at the bottom, which is a primary feature of this dock. It also omits other visible icons like VLC and the Ubuntu Software Center. The revision aims to create a more comprehensive description by including these missing details and to refine the functionality to encompass all its roles, including launching favorites, switching between running apps, and accessing the full application list.",\\
  "revised description": "A vertical dock is positioned on the left edge of the screen against a dark, semi-opaque background. It features a column of application icons, including Google Chrome, VLC media player, the LibreOffice suite, a file manager, and the Ubuntu Software Center. The LibreOffice Calc icon is highlighted with a gray background and a small orange dot, indicating it is the currently active application. Towards the bottom, separated by a thin line, are icons for the Trash and a grid of nine dots used to display all installed applications.",\\
  "revised functionality": "This dock serves as the primary application launcher and task manager for the Ubuntu operating system, providing quick access to favorite and running applications, as well as a gateway to view all installed programs." \\
  \}
\verb|```|

\end{longtable}
\twocolumn

\subsection{AutoGUI-v2 Task Generation Prompts}
The prompts used by AutoGUI-v2 to generate task are listed in Tab.~\ref{tab:group prompt 1}, Tab.~\ref{tab:group prompt 2}, Tab.~\ref{tab:region gnd prompt}, and Tab.~\ref{tab:region cap prompt}.
\onecolumn
\begin{longtable}{p{\dimexpr \textwidth - 2\tabcolsep}}

    \caption{The prompt used to instruct Gemini-2.5-Pro-Thinking to visually verify and refine groups of similar UI regions.} 
    \label{tab:group prompt 1} \\
    \toprule
    \textbf{Input Prompt Template} \\
    \midrule
    \endfirsthead
    
    \multicolumn{1}{c}{{\bfseries \tablename\ \thetable{} -- continued from previous page}} \\
    \toprule
    \endhead

    \midrule
    \multicolumn{1}{r}{{Continued on next page...}} \\
    \endfoot

    \bottomrule
    \endlastfoot

    
 <image> You are a GUI understanding expert. Your task is to verify and refine groups of visually similar UI regions.

Background:\\
We have identified a potential group of UI regions based on visual description similarity (using text embeddings of their visual appearance descriptions). However, text-based visual descriptions may not perfectly capture true visual similarity. Your role is to:\\
1. Verify if the initially grouped regions are truly visually similar\\
2. Check if any other candidate regions should be added to this group\\
3. Ensure the final group size is between 2-5 regions\\

Criteria for Valid Group Regions:\\
- Regions should be visually similar (similar icon type, similar appearance, similar color/shape)\\
- Regions should have different functionalities in their respective contexts\\
- Regions should be confusing for AI agents because they look alike but behave differently\\
- CRITICAL: Regions must NOT have overlapping bounding boxes (no parent-child or containment relationships)\\
  Check that no region's bbox is contained within another region's bbox\\
  If bbox A is inside bbox B, they cannot be in the same group\\

Initially Identified Group (based on visual description embeddings from Qwen3-Embedding):\\

\{initial\_group\_info\}

Other Candidate Regions (not in the initial group):\\
\{other\_candidates\_info\}

Your Task (Two-Stage Process):\\

Stage 1: Validate Initial Group Members\\
- Examine each region in the initial group\\
- Check if it truly meets all criteria (visually similar + different functionality + confusing + NO bbox overlap)\\
- Check for bbox overlaps: If any two regions have overlapping bboxes, keep only ONE of them\\
- Mark regions that don't fit as keep=false

Stage 2: Supplement from Other Candidates\\
- Examine all other candidate regions\\
- Check if any of them are visually similar to the group and should be added\\
- BEFORE adding: Verify the candidate's bbox does NOT overlap with ANY existing group member's bbox\\
- Only add candidates that pass ALL criteria requirement

Stage 3: Size Control\\
- After Stages 1 \& 2, count the final group size\\
- If size < 2: Mark the entire group as INVALID\\
- If size = 2-5: Keep all regions, mark as VALID\\
- If size > 5: Select the 5 most visually similar and confusing regions (ensuring no bbox overlaps), mark others as keep=false

IMPORTANT: Judging Logic\\
- Judge the group as VALID if at least 2 regions (after refinement) meet ALL criteria\\
- It's acceptable if some initial regions don't meet the criteria - just set their "keep" to false\\
- You MUST check all other candidates to see if they should be added\\
- Example: If initial group has 3 regions but only 1 fits, check other candidates. If you find 1+ qualifying candidates, add them and mark as VALID\\
- Only mark as INVALID if fewer than 2 regions can meet all criteria after checking ALL candidates

Output Format (JSON):\\
\{\\
  "valid": true/false,\\
  "rejection\_reason": "Explanation if invalid" (only if valid=false),\\
  "visual\_similarity\_description": "What makes these regions look similar" (only if valid=true),\\
  "kept\_region\_ids": [\\
    \{\\
      "region\_id": "1-0",\\
      "keep": true,  // false to remove from group\\
      "source": "initial\_group",  // or "added\_from\_candidates"\\
      "reason": "Brief reason for keeping or removing this region"\\
    \},\\
    \{\\
      "region\_id": "1-2",\\
      "keep": true,\\
      "source": "initial\_group",\\
      "reason": "Matches the visual pattern perfectly"\\
    \},\\
    \{\\
      "region\_id": "1-5",\\
      "keep": false,\\
      "source": "initial\_group",\\
      "reason": "Different visual style, doesn't match the group"\\
    \},\\
    \{\\
      "region\_id": "3-8",\\
      "keep": true,\\
      "source": "added\_from\_candidates",\\
      "reason": "Found in other candidates, matches the visual pattern"\\
    \}\\
  ],\\
  "final\_group\_size": 3,\\
  "adjustments\_made": "Removed 1 region from initial group (not visually similar), added 1 region from candidates"\\
\}

Requirements:\\
- Be strict: only approve groups that truly meet all criteria\\
- Visual similarity is critical - don't approve groups where regions just happen to have similar text\\
- Functionality differences must be clear and meaningful\\
- NO bbox overlaps allowed: Reject or remove any region whose bbox overlaps with another group member\\
- You MUST examine all other candidates, not just the initial group\\
- Final group size must be 2-5 regions (if >5, select the best 5 with no overlaps)

Now analyze the screenshot and evaluate this group:
 \\
    \midrule

    \textbf{Response Example}\\
    \midrule

<think>\\
Alright, here's the deal. I've been asked to review a bunch of UI elements and group them together, making sure they're visually similar, don't overlap, and could potentially trip up an AI. The final group needs to have between 2 and 5 elements.\\
First, I examined the starting list. I noticed the menu bar, tab bar, application title bar, and settings sidebar. The title bar and settings sidebar are going to have to go. The title bar is light gray, unlike the others, and the settings sidebar is vertical and quite different. The menu bar and the tab bar, however, are both dark horizontal bars, so they stay.\\
Then, I went through the rest of the list. Found the status bar at the bottom, and the search bar, which are both dark and horizontal, so they're added to the mix. The OS-level top bar is also added; it looks a lot like the app's top bars. This leaves me with the menu bar, tab bar, status bar, search bar, and OS top bar. They're all horizontal, dark, and they all have some sort of light-colored text or icons.\\
The final grouping is complete. I've noted the reasons I kept the original elements, and why I added the other candidates. They’re all horizontal bars and look very similar. This should be a good group, and hopefully, will cause an AI some confusion. I'm ready to write up the final report with the details.\\
</think>\\
\verb|```|json\\
\{\\
  "valid": true,\\
  "rejection\_reason": null,\\
  "visual\_similarity\_description": "All regions are dark, horizontal bars with light-colored text or icons, serving as toolbars, status bars, or navigation elements at different levels of the UI (OS and application).",\\
  "kept\_region\_ids": [\\
    \{\\
      "region\_id": "2-5",\\
      "keep": true,\\      
      "source": "initial\_group",\\      
      "reason": "A dark horizontal menu bar with light text, fits the group's visual theme."\\    
    \},\\    
    \{\\      
      "region\_id": "4-7",\\      
      "keep": false,\\      
      "source": "initial\_group",\\      
      "reason": "This is a vertical navigation list, not a horizontal bar. Its orientation and structure are different from the group."\\
    \},\\    
    \{\\      
      "region\_id": "2-4",\\      
      "keep": false,\\      
      "source": "initial\_group",\\      
      "reason": "This is a light gray title bar with dark text, which is visually distinct from the dark-themed bars in the group."\\    
    \},\\    
    \{\\      
      "region\_id": "3-8",\\      
      "keep": true,\\      
      "source": "initial\_group",\\      
      "reason": "A dark horizontal tab bar with light text, fitting the group's visual theme."\\    
    \},\\    
    \{\\      
      "region\_id": "2-9",\\      
      "keep": true,\\      
      "source": "added\_from\_candidates",\\      
      "reason": "This is the application's bottom status bar, a dark horizontal bar with light text/icons that matches the group's visual theme."\\    
    \},\\    
    \{\\      
      "region\_id": "4-4",\\      
      "keep": true,\\      
      "source": "added\_from\_candidates",\\      
      "reason": "This is a dark horizontal search bar with light text, which fits the visual pattern of the group."\\    
    \},\\    
    \{\\      
      "region\_id": "1-0",\\      
      "keep": true,\\      
      "source": "added\_from\_candidates",\\      
      "reason": "This is the OS-level top bar, which is a dark horizontal bar with light text, matching the group's visual characteristics."\\    
    \}\\  
  ],\\  
  "final\_group\_size": 5,\\  
  "adjustments\_made": "Removed 2 regions from the initial group due to different visual styles (light theme, vertical orientation). Added 3 visually similar regions from the candidates (application status bar, search bar, and OS top bar)."\\
\}\\
\verb|```| \\
\end{longtable}
\twocolumn
\onecolumn
\begin{longtable}{p{\dimexpr \textwidth - 2\tabcolsep}}

    \caption{The prompt used to instruct Gemini-2.5-Pro-Thinking to select the optimal subset of regions from oversized or merged groups.} 
    \label{tab:group prompt 2} \\
    \toprule
    \textbf{Input Prompt Template} \\
    \midrule
    \endfirsthead
    
    \multicolumn{1}{c}{{\bfseries \tablename\ \thetable{} -- continued from previous page}} \\
    \toprule
    \endhead

    \midrule
    \multicolumn{1}{r}{{Continued on next page...}} \\
    \endfoot

    \bottomrule
    \endlastfoot

    
 <image> You are a GUI understanding expert. Your task is to select the BEST 2-5 regions from a large group of visually similar UI regions.

Background:\\
We have identified a group of UI regions that are visually similar and functionally different. However, the group might have MORE than 5 regions, which is too many for a multiple-choice question. Your role is to:\\
1. Select the 2-5 MOST representative and confusing regions\\
2. Ensure selected regions have NO bbox overlaps\\
3. Maximize visual similarity while maintaining functional diversity

Group Information:\\
\{group\_info\}

Selection Criteria (in order of priority):\\
1. Visual Similarity: Select regions that look MOST similar to each other\\
2. Functional Diversity: Ensure selected regions have clearly DIFFERENT functionalities\\
3. Confusion Potential: Prioritize regions that would be MOST confusing for AI agents\\
4. NO Bbox Overlaps: Selected regions must NOT have overlapping bounding boxes\\
5. Optimal Size: Select 2-5 regions (prefer 3-4 if possible for better question quality)

Your Task:\\
1. Analyze all regions in the group (ONLY the regions listed in "Group Information" above)\\
2. Identify the core "visual pattern" (e.g., all are blue icons, all are text buttons)\\
3. Select 2-5 regions that BEST represent this pattern\\
4. CRITICAL: You MUST ONLY select regions from the group list provided above. Do NOT select any region\_id that is NOT in the "Group Information" section.\\
5. Ensure NO two selected regions have overlapping bboxes\\
6. Provide clear reasoning for your selection

Output Format (JSON):\\
\{\\
  "selected\_region\_ids": [\\
    \{\\
      "region\_id": "1-0",\\
      "selection\_reason": "Why this region was selected"\\
    \},\\
    \{\\
      "region\_id": "1-2",\\
      "selection\_reason": "Why this region was selected"\\
    \},\\
    // ... 2-5 regions total\\
  ],\\
  "excluded\_region\_ids": [\\
    \{\\
      "region\_id": "1-3",\\
      "exclusion\_reason": "Why this region was excluded"\\
    \},\\
    // ... other excluded regions\\
  ],\\
  "visual\_pattern": "Description of the common visual pattern",\\
  "final\_count": 3,\\
  "selection\_summary": "Brief summary of selection strategy"\\
\}

Requirements:\\
- CRITICAL: You MUST ONLY select region\_ids from the "Group Information" list above. Any region\_id NOT in that list will be rejected.\\
- You MUST select between 2-5 regions (no more, no less)\\
- Selected regions MUST have NO bbox overlaps\\
- Prioritize regions with highest visual similarity\\
- Ensure functional diversity among selected regions\\
- Provide clear reasoning for each selection/exclusion

Now analyze the screenshot and select the best regions:
 \\
    \midrule

    \textbf{Response Example}\\
    \midrule

<think>\\
Okay, let's break this down. The task is to pick 2-5 visually similar UI elements from a provided list, and given my expertise, I need to choose wisely to maximize the potential for interesting analysis.\\
First, I see the core pattern: all four elements are dark gray bars. They act as containers for information and/or navigation and are located at the edges of the screen or a window. There are horizontal (top bar, status bar) and vertical (dock, activity bar) bars, so I need to account for both orientations. I recognize the difference between OS-level and application-level components, which adds another layer to the analysis.\\
The list includes: Ubuntu Top Bar, VS Code Activity Bar, Ubuntu Dock, and VS Code Status Bar.\\
Now, let's look at the confusing pairs. The Ubuntu Dock (OS-level, vertical) and the VS Code Activity Bar (app-level, vertical) are immediate red flags. They are visually very similar – icon-based vertical bars on the left – but have totally different purposes. Then there's the Ubuntu Top Bar (OS-level, horizontal) and the VS Code Status Bar (app-level, horizontal). Again, both display status information but at different levels.\\
My strategy needs to highlight both visual similarity AND functional diversity to create the most interesting and challenging question.\\
Considering the boundaries and similarities, I should aim for maximum coverage. I'm going to select all four elements.\\
*   **1-0 (Ubuntu Top Bar):** OS, horizontal, top. That's a strong start.\\
*   **1-1 (Ubuntu Dock):** OS, vertical, left. This complements the top bar.\\
*   **3-13 (VS Code Activity Bar):** App, vertical, left. This is key for the core confusion around OS vs. app chrome.\\
*   **2-9 (VS Code Status Bar):** App, horizontal, bottom. This completes the set.\\
By including all four, I have the key axes of consideration - OS vs App, and vertical vs horizontal - perfectly covered. The similarities are in the dark gray bar aesthetic, and the differences are in their exact location, function, and target (OS or App). This should result in a great selection for the user! The final count of selected regions is therefore four. I'm confident in this selection.\\
</think>\\
\verb|```|json\\
\{\\  
  "selected\_region\_ids": [\\    
    \{\\      
      "region\_id": "1-0",\\      
      "selection\_reason": "This is the OS-level top bar. It's a horizontal, dark gray bar, visually similar to the application's status bar (2-9) but functionally distinct (system-wide info vs. project-specific status)."\\ 
    \},\\    
    \{\\      
      "region\_id": "3-13",\\      
      "selection\_reason": "This is the application's internal 'Activity Bar'. It's a vertical, dark gray icon bar that is highly confusable with the OS-level dock (1-1) due to its appearance and position, but it controls navigation within the app, not between apps."\\    
    \},\\    
    \{\\      
      "region\_id": "1-1",\\      
      "selection\_reason": "This is the OS-level dock. It's a vertical, dark gray icon bar for launching and switching applications. Its visual similarity to the VS Code Activity Bar (3-13) creates significant potential for confusion."\\    
    \},\\    
    \{\\      
      "region\_id": "2-9",\\      
      "selection\_reason": "This is the application's status bar at the bottom. It's a horizontal, dark gray bar that mirrors the OS top bar (1-0) in layout but serves a different, application-specific purpose (displaying project status)."\\    
    \}\\  
  ],\\  
  "excluded\_region\_ids": [],\\  
  "visual\_pattern": "The common visual pattern is a dark gray bar, either horizontal or vertical, positioned at the edge of the screen or application window, which serves as a container for icons and text related to navigation or status.",\\  
  "final\_count": 4,\\  
  "selection\_summary": "The selection includes all four available regions as they perfectly represent the theme of visually similar but functionally distinct UI 'chrome' elements. The group creates a comprehensive comparison between OS-level bars (top bar, dock) and application-level bars (activity bar, status bar), and also between horizontal and vertical layouts. This maximizes both visual similarity and functional diversity, creating a challenging and informative set."\\
\}\\
\verb|```| \\
\end{longtable}
\twocolumn
\onecolumn
\begin{longtable}{p{\dimexpr \textwidth - 2\tabcolsep}}

    \caption{The prompt used to instruct Gemini-2.5-Pro-Thinking to construct functionality-oriented grounding task.} 
    \label{tab:region gnd prompt} \\
    \toprule
    \textbf{Input Prompt Template} \\
    \midrule
    \endfirsthead
    
    \multicolumn{1}{c}{{\bfseries \tablename\ \thetable{} -- continued from previous page}} \\
    \toprule
    \endhead

    \midrule
    \multicolumn{1}{r}{{Continued on next page...}} \\
    \endfoot

    \bottomrule
    \endlastfoot

    
Based on the following visually similar regions with different functionalities, generate a question to test if an agent can predict the outcome or purpose of interacting with them.\\

Region Information:\\
\{region\_info\}

Task:\\
Generate a question asking the agent to predict what will happen when interacting with a specific region, or what purpose/goal the interaction serves.

Question Format Guidelines:\\
Focus on PREDICTION and OUTCOME, such as:\\
- "If you want to [achieve specific goal], which region should you click?"\\
- "Which region will [result/outcome] when clicked?"\\
- "To [specific purpose/intention], which region would you interact with?"\\
- "Clicking which region will lead to [specific result/interface]?"

Output Format (JSON):\\
\{\\
  "question": "If you want to [specific goal/purpose], which region should you click?",\\
  "options": [\\
    \{\\
      "label": "A",\\
      "region\_id": "1-0"\\
    \},\\
    \{\\
      "label": "B",\\
      "region\_id": "1-2"\\
    \},\\
    // ... include ALL regions from the group as options\\
  ],\\
  "correct\_answer": "A",  // The label of correct option\\
  "explanation": "Why clicking this region will achieve the goal/produce the expected result"\\
\}\\

Requirements:\\
- CRITICAL: You MUST include ALL regions from the group information as options (do not select a subset)\\
- Each option only needs "label" and "region\_id" - do NOT copy bbox or description from the input (we already have them)\\
- Question must focus on PREDICTION: what will happen, what goal will be achieved, what result will occur\\
- Avoid questions like "Which region is for X?" - instead ask "To achieve X, which region should you use?"\\
- Options should be ordered randomly (not by position)\\
- All options except the correct one should be plausible distractors\\
- Explanation should clearly describe the predicted outcome/result\\

Please generate the question now:
 \\
    \midrule

    \textbf{Response Example}\\
    \midrule

<think>\\
**Formulating a Predictive Multiple-Choice Question**\\
Okay, so I've got this JSON data outlining different system regions, and my task is to create a multiple-choice question that tests the user's understanding of their functionality. Let's break this down.\\
First, I need to really *understand* the data. I see a group with two regions: 2-11, which looks like a clock and notification indicator, and 2-12, a cluster of system icons. Both are on the top bar, visually similar (small, white icons/text), but they perform different tasks. This is perfect for a question that focuses on function over pure appearance.\\
Brainstorming, I'm aiming for questions that predict the outcome of an interaction. The core of this is about the *action* someone would take to achieve a specific *goal*. I thought of a few ideas:\\
*   One focusing on notifications/calendar (Region 2-11)\\
*   Another on system volume/Wi-Fi (Region 2-12)\\
*   And a slightly more abstract one about general system settings.\\
I'm leaning towards the system volume/Wi-Fi example because it's specific and directly assesses the functionality of Region 2-12. So, I settled on "Clicking which region will allow you to adjust the system volume and manage network connections?".\\
Now, let's build the question. I need to make sure I include all regions from the group, randomize the option order, and define a clear explanation.\\
The answer is Region 2-12 (option A in the randomly ordered choices), as it directly relates to managing system settings. The explanation has to justify the correct choice by stating that it opens the system control menu with the specific settings needed.\\
Finally, I put it all together in the structured JSON format: the question, randomized options with their labels and region IDs, the correct answer's label, and a solid explanation. Done. It's a question focused on prediction with clear links to the functionality of the system regions. The question is testing understanding, it's not simply memorization.\\
</think>\\
\verb|```|json\\
\{\\  
  "question": "If you want to adjust the system volume or manage your network connections, which region should you click?",\\  
  "options": [\\    
    \{\\      
      "label": "A",\\      
      "region\_id": "2-11"\\    
    \},\\    
    \{\\      
      "label": "B",\\      
      "region\_id": "2-12"\\    
    \}\\  
  ],\\  
  "correct\_answer": "B",\\  
  "explanation": "Clicking this region will open the system control menu, which contains controls for managing network connections, adjusting audio volume, and accessing power settings."\\
\}\\
\verb|```| \\
\end{longtable}
\twocolumn
\onecolumn
\begin{longtable}{p{\dimexpr \textwidth - 2\tabcolsep}}

    \caption{The prompt used to instruct Gemini-2.5-Pro-Thinking to construct functionality-oriented captioning task.} 
    \label{tab:region cap prompt} \\
    \toprule
    \textbf{Input Prompt Template} \\
    \midrule
    \endfirsthead
    
    \multicolumn{1}{c}{{\bfseries \tablename\ \thetable{} -- continued from previous page}} \\
    \toprule
    \endhead

    \midrule
    \multicolumn{1}{r}{{Continued on next page...}} \\
    \endfoot

    \bottomrule
    \endlastfoot

    
 <image> You are an expert in GUI analysis. I will show you a GUI screenshot with ONE UI region highlighted by a red bounding box.

This region belongs to a group of visually similar regions with different functionalities:\\
\{group\_regions\_info\}

Target region (the one with red box):\\
\{target\_region\_info\}

Task:\\
Create a question asking about what will happen or what goal will be achieved when clicking the circled region.\\
The options should describe the outcomes/purposes of ALL regions in this group (including the target region).

Question Format:\\
Focus on PREDICTION and OUTCOME:\\
- "If you click the circled region, what will happen?"\\
- "What is the expected result of clicking the circled region?"\\
- "What goal can be achieved by clicking the circled region?"

Output Format (JSON):\\
\{\\
  "question": "If you click the circled region, what will happen?",\\
  "options": [\\
    \{\\
      "label": "A",\\
      "region\_id": "1-0",\\
      "functionality": "Description of expected outcome/result (rephrased from functionality)"\\
    \},\\
    \{\\
      "label": "B",\\
      "region\_id": "1-2",\\
      "functionality": "Description of expected outcome/result (rephrased from functionality)"\\
    \},\\
    \{\\
      "label": "C",\\
      "region\_id": "2-1",\\
      "functionality": "Description of expected outcome/result (rephrased from functionality)"\\
    \}\\
    // ... include all group regions as options\\
  ],\\
  "correct\_answer": "A",  // The label of the option containing target region's outcome\\
  "explanation": "Why this outcome/result will occur when clicking the circled region, and why other outcomes (from similar-looking regions) are incorrect"\\
\}\\

Requirements:\\
- Question must focus on PREDICTION: what will happen, what result will occur, what goal will be achieved\\
- Rephrase each functionality as an outcome/result (e.g., "Save document" → "The document will be saved")\\
- Include ALL group regions' outcomes as options (the group may contain 2 or more regions)\\
- One option must be the target region's outcome (the circled one)\\
- Options should be shuffled randomly (not by spatial position)\\
- Each option must include the region\_id to identify which region it corresponds to\\
- Explanation should describe the predicted outcome and highlight how to distinguish the target from other visually similar regions\\

Please generate the question now:
 \\
    \midrule

    \textbf{Response Example}\\
    \midrule

<think>\\
**Crafting the Perfect UI Question**\\
Okay, so I've been given a screenshot and a task: create a multiple-choice question about the Ubuntu Dock, specifically predicting the *outcome* of clicking an icon within that dock. This is the thought process I'm following:\\
First, I'm carefully analyzing the input. I have a screenshot of Ubuntu with VS Code open. The key here is the circled region – the Ubuntu Dock on the left.  Crucially, I also have a *group* of four visually similar regions – the top bar, the VS Code Activity Bar, the Ubuntu Dock (my target), and the VS Code Status Bar – each with distinct functionality.\\
Next, I need to translate those functionalities into concrete, testable *outcomes*. For the top bar, it's about system-level access and application management. The VS Code Activity Bar is about switching between internal editor views. The *target*, the Ubuntu Dock, is all about launching or switching applications. Finally, the VS Code Status Bar provides project state information.\\
Now, formulating the question itself. I've decided on "What goal can be achieved by clicking an icon within the circled region?" because it's precise about the *action* (clicking) and the *outcome* (a goal). It's more about the function the region has, rather than the region itself.\\
The heart of the question is the answer *options*. These have to accurately represent the outcomes of *all* the regions in the group, including the Ubuntu Dock. So, I'm crafting each option to concisely capture the expected result of clicking each relevant UI element.  I'm also careful to include each option with its appropriate `region\_id` and ensuring that the list is shuffled for randomness.\\
Finally, the *explanation*. This is where the learning happens! I'll start by clearly identifying the correct answer (the Ubuntu Dock). Then, I'll systematically *contrast* the other options, explaining what makes them *different*. This is key – the explanation needs to highlight the nuances between each region's purpose, emphasizing their *unique* functionalities within the broader UI context. I'll make sure to highlight the difference in *scope*; system-wide function versus application function.\\
And lastly, I assemble the JSON output, double-checking all the details: question wording, option mapping, correct answer, and the insightful explanation. I'm aiming for clarity, conciseness, and, above all, an educational value in this question. That's the goal!\\
</think>\\
\verb|```|json\\
\{\\  
  "question": "What is the expected result of clicking an icon within the circled region?",\\  
  "options": [\\    
    \{\\      
      "label": "A",\\      
      "region\_id": "3-13",\\      
      "functionality": "You will switch between different tool panels within the Visual Studio Code application, such as the file explorer, search, or source control."\\    
    \},\\    
    \{\\      
      "label": "B",\\      
      "region\_id": "1-0",\\      
      "functionality": "You will access system-level functions, such as the Activities overview for managing open windows, the calendar, or system settings."\\    
    \},\\    
    \{\\      
      "label": "C",\\      
      "region\_id": "1-1",\\      
      "functionality": "You will launch a favorite application or switch to an application that is already running on the desktop."\\    
    \},\\    
    \{\\      
      "label": "D",\\      
      "region\_id": "2-9",\\      
      "functionality": "You will view information about the current coding project, such as the Git branch, synchronization status, or project-wide errors and warnings."\\    
    \}\\  
  ],\\  
  "correct\_answer": "C",\\  
  "explanation": "The circled region is the Ubuntu operating system's dock. Its purpose is to launch applications pinned to it or to switch focus to applications that are already open. It is distinct from the other vertical bar with icons (Region 3-13), which is the Activity Bar inside the Visual Studio Code application and is used to navigate between different views within the editor itself. The horizontal bar at the very top (Region 1-0) is for system-level controls, and the horizontal bar at the bottom of the VS Code window (Region 2-9) is the status bar, which shows project-specific information."\\
\}\\
\verb|```| \\
\end{longtable}
\twocolumn

\subsection{AutoGUI-v2 Evaluation Prompts}
The prompts used for evaluating VLMs on the AutoGUI-v2 functionality, description, and intent-based element grounding tasks are shown in Tab.~\ref{tab:func elemgnd prompts}, Tab.~\ref{tab:desc elemgnd prompts}, Tab.~\ref{tab:intent elemgnd prompts}.

The prompt used for evaluation on the AutoGUI-v2 functionality-based region grounding tasks is shown in Tab.~\ref{tab:func elemgnd prompts}.

The prompt used to evaluate VLMs on the functionality-based element captioning task is shown in Tab.~\ref{tab:func elemcap prompt}. Note that the bounding box of a target element/region is drawn in the screenshot as a red rectangle.

\begin{table*}[th]
\centering
\caption{The evaluation prompts used in the functionality-based element/region grounding task. The placeholder `task\_type' is replaced with `element' in the element grounding task and `region' in the region grounding task.}
\label{tab:func elemgnd prompts}
\resizebox{\textwidth}{!}{%
\begin{tabular}{@{}cl@{}}
\toprule
\textbf{Model}                                                                 & \multicolumn{1}{c}{\textbf{Prompt}}                                                                                                                                                                                                                                                                                                                                                                                                                                                                                                                                                                                 \\ \midrule
Gemini-2.5-Pro-Thinking                                                        & \begin{tabular}[c]{@{}l@{}}You are a GUI expert. Given a screenshot and a question about locating a specific UI \{target\_type\},\\ you need to identify the bounding box of the target \{target\_type\}, which should be {[}ymin, xmin, ymax, xmax{]}\\ normalized to 0-1000. Note that the X-axis runs horizontally from left (0) to right (999),\\ and the Y-axis runs vertically from top (0) to bottom (999).\\ \\ Description: \{question\}\\ Now analyze the screenshot and provide the bounding box for the target element:\end{tabular}                                                                                         \\
\rowcolor[HTML]{EFEFEF} 
Claude-Sonnet-4.5                                                              & \begin{tabular}[c]{@{}l@{}}You are a GUI expert. Given a screenshot and a question about locating a specific UI \{target\_type\},\\ you need to identify the bounding box of the target element, which should be {[}xmin, ymin, xmax, ymax{]}.\\ Note that the X-axis runs horizontally from left (0) to right (999),\\ and the Y-axis runs vertically from top (0) to bottom (999).\\ \\ Description: \{question\}\\ \\ Output format:\\ Box: {[}xmin, ymin, xmax, ymax{]}\\ Now analyze the screenshot and provide the bounding box for the target element:\end{tabular}                                                      \\
UI-Tars-1.5                                                                    & \begin{tabular}[c]{@{}l@{}}You are a GUI agent. You are given a task and your action history, with screenshots.\\ You need to perform the next action to complete the task.\\ \\ \#\# Output Format\\ \\ Action: ...\\ \\ \\ \#\# Action Space\\ click(point=`<point>x1 y1</point>')\\ \\ \#\# User Instruction\\ \{question\}\end{tabular}                                                                                                                                                      \\
\rowcolor[HTML]{EFEFEF} 
OS-Atlas-Base-7B                                                               & In this UI screenshot, what is the position of the \{target\_type\} corresponding to the command "\{question\}" (with bbox)?                                                                                                                                                                                                                                                                                                                                                                                                                                                                                                 \\
UGround-V1-7B                                                                  & \begin{tabular}[c]{@{}l@{}}Your task is to help the user identify the precise coordinates (x, y) of a specific area/element/object on the screen based on a description.\\ \\ - Your response should aim to point to the center or a representative point within the described area/element/object as accurately as possible.\\ - If the description is unclear or ambiguous, infer the most relevant area or element based on its likely context or purpose.\\ - Your answer should be a single string (x, y) corresponding to the point of the interest.\\ \\ Description: \{instruction\}\\ Answer:\end{tabular} \\
\rowcolor[HTML]{EFEFEF} 
\begin{tabular}[c]{@{}c@{}}O3, GPT-5, Step-3,\\ Qwen3-VL, GLM-4.5V\end{tabular} & \begin{tabular}[c]{@{}l@{}}You are a GUI expert. Given a screenshot and a question about locating a specific UI \{target\_type\},\\ you need to identify the bounding box of the target element, which should be {[}xmin, ymin, xmax, ymax{]}\\ normalized to 0-1000. Note that the X-axis runs horizontally from left (0) to right (999),\\ and the Y-axis runs vertically from top (0) to bottom (999).\\ \\ Description: \{question\}\\ \\ Output format:\\ Box: [xmin, ymin, xmax, ymax]\\ Now analyze the screenshot and provide the bounding box for the target element:\end{tabular}                         \\ \bottomrule
\end{tabular}%
}
\end{table*}

\begin{table*}[th]
\centering
\caption{The evaluation prompts used in the appearance-description-based element grounding task.}
\label{tab:desc elemgnd prompts}
\resizebox{\textwidth}{!}{%
\begin{tabular}{@{}cl@{}}
\toprule
\textbf{Model}                                                                 & \multicolumn{1}{c}{\textbf{Prompt}}                                                                                                                                                                                                                                                                                                                                                                                                                                                      \\ \midrule
Gemini-2.5-Pro-Thinking                                                        & \begin{tabular}[c]{@{}l@{}}You are a GUI expert. Given a screenshot and a description of a specific UI element,\\ you need to identify the bounding box of the target element, which should be {[}ymin, xmin, ymax, xmax{]}\\ normalized to 0-1000. Note that the X-axis runs horizontally from left (0) to right (999),\\ and the Y-axis runs vertically from top (0) to bottom (999).\\ \\ Question: \{question\}\\ Now analyze the screenshot and provide the bounding box for the target element:\end{tabular}                                                                                         \\
\rowcolor[HTML]{EFEFEF} 
Claude-Sonnet-4.5                                                              & \begin{tabular}[c]{@{}l@{}}You are a GUI expert. Given a screenshot and a description of a specific UI element,\\ you need to identify the bounding box of the target element, which should be {[}xmin, ymin, xmax, ymax{]}.\\ Note that the X-axis runs horizontally from left (0) to right (999),\\ and the Y-axis runs vertically from top (0) to bottom (999).\\ \\ Question: \{question\}\\ \\ Output format:\\ Box: {[}xmin, ymin, xmax, ymax{]}\\ Now analyze the screenshot and provide the bounding box for the target element:\end{tabular}                                                      \\
UI-Tars-1.5                                                                    & \begin{tabular}[c]{@{}l@{}}You are a GUI agent. You are given a task and your action history, with screenshots.\\ You need to perform the next action to complete the task.\\ \\ \#\# Output Format\\ \\ Action: ...\\ \\ \\ \#\# Action Space\\ click(point=`<point>x1 y1</point>')\\ \\ \#\# User Instruction\\ \{question\}\end{tabular}                                                                                                                                                      \\
\rowcolor[HTML]{EFEFEF} 
OS-Atlas-Base-7B                                                               & In this UI screenshot, what is the position of the element corresponding to the command "\{question\}" (with bbox)?                                                                                                                                                                                                                                                                                                                                                                                                                                                                                                 \\
UGround-V1-7B                                                                  & \begin{tabular}[c]{@{}l@{}}Your task is to help the user identify the precise coordinates (x, y) of a specific area/element/object on the screen based on a description.\\ \\ - Your response should aim to point to the center or a representative point within the described area/element/object as accurately as possible.\\ - If the description is unclear or ambiguous, infer the most relevant area or element based on its likely context or purpose.\\ - Your answer should be a single string (x, y) corresponding to the point of the interest.\\ \\ Description: \{instruction\}\\ Answer:\end{tabular} \\
\rowcolor[HTML]{EFEFEF} 
\begin{tabular}[c]{@{}c@{}}O3, GPT-5, Step-3,\\ Qwen3-VL, GLM-4.5V\end{tabular} & \begin{tabular}[c]{@{}l@{}}You are a GUI expert. Given a screenshot and a description of a specific UI element,\\ you need to identify the bounding box of the target element, which should be {[}xmin, ymin, xmax, ymax{]}\\ normalized to 0-1000. Note that the X-axis runs horizontally from left (0) to right (999),\\ and the Y-axis runs vertically from top (0) to bottom (999).\\ \\ Question: \{question\}\\ \\ Output format:\\ Box: [xmin, ymin, xmax, ymax]\\ Now analyze the screenshot and provide the bounding box for the target element:\end{tabular}                         \\ \bottomrule
\end{tabular}%
}
\end{table*}
\begin{table*}[th]
\centering
\caption{The evaluation prompts used in the action-intent-based element grounding task.}
\label{tab:intent elemgnd prompts}
\resizebox{\textwidth}{!}{%
\begin{tabular}{@{}cl@{}}
\toprule
\textbf{Model}                                                                 & \multicolumn{1}{c}{\textbf{Prompt}}                                                                                                                                                                                                                                                                                                                                                                                                                                                      \\ \midrule
Gemini-2.5-Pro-Thinking                                                        & \begin{tabular}[c]{@{}l@{}}You are a GUI expert. Given a screenshot and an action intent about interacting with a specific UI element,\\ you need to identify the bounding box of the target element, which should be {[}ymin, xmin, ymax, xmax{]}\\ normalized to 0-1000. Note that the X-axis runs horizontally from left (0) to right (999),\\ and the Y-axis runs vertically from top (0) to bottom (999).\\ \\ Action Intent: \{question\}\\ Now analyze the screenshot and provide the bounding box for the target element:\end{tabular}                                                                                         \\
\rowcolor[HTML]{EFEFEF} 
Claude-Sonnet-4.5                                                              & \begin{tabular}[c]{@{}l@{}}You are a GUI expert. Given a screenshot and an action intent about interacting with a specific UI element,\\ you need to identify the bounding box of the target element, which should be {[}xmin, ymin, xmax, ymax{]}.\\ Note that the X-axis runs horizontally from left (0) to right (999),\\ and the Y-axis runs vertically from top (0) to bottom (999).\\ \\ Action Intent: \{question\}\\ \\ Output format:\\ Box: {[}xmin, ymin, xmax, ymax{]}\\ Now analyze the screenshot and provide the bounding box for the target element:\end{tabular}                                                      \\
UI-Tars-1.5                                                                    & \begin{tabular}[c]{@{}l@{}}You are a GUI agent. You are given a task and your action history, with screenshots.\\ You need to perform the next action to complete the task.\\ \\ \#\# Output Format\\ \\ Action: ...\\ \\ \\ \#\# Action Space\\ click(point=`<point>x1 y1</point>')\\ \\ \#\# User Instruction\\ \{question\}\end{tabular}                                                                                                                                                      \\
\rowcolor[HTML]{EFEFEF} 
OS-Atlas-Base-7B                                                               & In this UI screenshot, what is the position of the element corresponding to the command "\{question\}" (with bbox)?                                                                                                                                                                                                                                                                                                                                                                                                                                                                                                 \\
UGround-V1-7B                                                                  & \begin{tabular}[c]{@{}l@{}}Your task is to help the user identify the precise coordinates (x, y) of a specific area/element/object on the screen based on a description.\\ \\ - Your response should aim to point to the center or a representative point within the described area/element/object as accurately as possible.\\ - If the description is unclear or ambiguous, infer the most relevant area or element based on its likely context or purpose.\\ - Your answer should be a single string (x, y) corresponding to the point of the interest.\\ \\ Description: \{instruction\}\\ Answer:\end{tabular} \\
\rowcolor[HTML]{EFEFEF} 
\begin{tabular}[c]{@{}c@{}}O3, GPT-5, Step-3,\\ Qwen3-VL, GLM-4.5V\end{tabular} & \begin{tabular}[c]{@{}l@{}}You are a GUI expert. Given a screenshot and an action intent about interacting with a specific UI element,\\ you need to identify the bounding box of the target element, which should be {[}xmin, ymin, xmax, ymax{]}\\ normalized to 0-1000. Note that the X-axis runs horizontally from left (0) to right (999),\\ and the Y-axis runs vertically from top (0) to bottom (999).\\ \\ Action Intent: \{question\}\\ \\ Output format:\\ Box: [xmin, ymin, xmax, ymax]\\ Now analyze the screenshot and provide the bounding box for the target element:\end{tabular}                         \\ \bottomrule
\end{tabular}%
}
\end{table*}
\begin{table*}[tp]

\centering
\caption{The evaluation prompt used in the functionality-based element captioning task cast into multi-choice task.}
\label{tab:func elemcap prompt}
\begin{tabular}{|l|}
\hline
\begin{tabular}[c]{@{}p{1.0\linewidth}@{}}

You are a GUI expert. Read the question and options, analyze the given GUI screenshot, and then choose the single best answer.\\ \\ Answer with a JSON object only:\\ \{\{"answer": "A/B/C/D/E"\}\}\\ \\ Question:\\ \{question\}\\ \\ Options:\\ \{options\_block\}\\ Now provide your answer:

\end{tabular} \\ \hline
\end{tabular}
\end{table*}


\section{Societal Impact}
\label{sec:supp:social impact}

The advancement of VLM-based GUI agents enables significant accessibility benefits but introduces dual-use risks regarding labor displacement and digital security. We address these dimensions below:

\paragraph{Labor and Surveillance.} While AutoGUI-v2 aims to reduce cognitive load via automation, robust GUI agents could theoretically facilitate workplace surveillance or displace roles in data entry and QA. We argue that the ethical deployment of such agents requires a "human-in-the-loop" paradigm, where the agent functions as an augmenting copilot rather than an autonomous replacement.

\paragraph{Data Ethics and Bias.} Our dataset is constructed from open-source repositories and public interfaces, rigorously scrubbed for offensive content. We explicitly acknowledge a distributional bias toward English-language, Western-centric design patterns. Consequently, agents evaluated solely on this benchmark may lack cross-cultural generalization, and we caution against assuming universal applicability without further diverse testing.

\section{Limitations and Future Work}
AutoGUI-v2 provides a semi-automated GUI functional region discovery pipeline and a comprehensive functionality understanding benchmark. Nevertheless, AutoGUI-v2 faces the following limitations:
\begin{itemize}
    \item \textbf{Full automation is not yet achieved.} The AutoGUI-v2 annotation system requires human annotators to correct the bounding boxes of the detected functional regions. This human-in-the-loop annotation procedure restricts the scaling potential of AutoGUI-v2. Future work can explore a reliable self-verification that substitutes human labor with competent model assistance.
    \item \textbf{Lack of task relevance.} The AutoGUI-v2 possesses massive functionality and appearance descriptions for the detected functional regions. However, the functionality descriptions are bound with a single-step instruction, instead of a long task sequence. This means that the interaction outcome involves only the outcome of the next step, not more than two steps.
    \item \textbf{The relevance between functionality understanding and planning is needed.} Although a comprehensive analysis of functionality understanding ability is presented in the main script, how this ability correlates with ultimate planning ability in multi-platform interaction scenarios is not analyzed in this work. We will explore this essential topic in the future by investigating how functionality understanding is used in GUI agents' chain-of-thought reasoning and affects planning performance.
\end{itemize}
\label{sec:supp:limitations}


\end{document}